\documentclass[journal]{IEEEtran}
\usepackage{amsmath,amsfonts}
\usepackage{algorithmic}
\usepackage{array}
\usepackage{textcomp}
\usepackage{stfloats}
\usepackage{url}
\usepackage{verbatim}
\usepackage{graphicx}
\def\BibTeX{{\rm B\kern-.05em{\sc i\kern-.025em b}\kern-.08em
    T\kern-.1667em\lower.7ex\hbox{E}\kern-.125emX}}
\usepackage{balance}
\usepackage[capitalise]{cleveref}
\usepackage{subcaption}
\usepackage{xcolor}
\usepackage{bm}
\usepackage[noadjust]{cite}

\begin{document}

\title{Genetic Programming for \\Explainable Manifold Learning}

\author{

    Ben~Cravens,
    Andrew~Lensen,~\IEEEmembership{Member,~IEEE,}
    Paula~Maddigan,
	Bing~Xue,~\IEEEmembership{Fellow,~IEEE}\\%
   
	\thanks{This work was supported by the University Research Fund at Te Herenga Waka--Victoria University of Wellington under grant number 410128/4223.}%
	\thanks{The authors are with the Centre for Data Science and Artificial Intelligence, and School of Engineering and Computer Science;  Victoria University of Wellington; Wellington 6140; New Zealand (e-mail: ben.conor.reese.cravens@gmail.com; andrew.lensen@ecs.vuw.ac.nz; paula.maddigan@ecs.vuw.ac.nz; bing.xue@ecs.vuw.ac.nz).}%
    \thanks{\copyright 2025 IEEE.  Personal use of this material is permitted.  Permission from IEEE must be obtained for all other uses, in any current or future media, including reprinting/republishing this material for advertising or promotional purposes, creating new collective works, for resale or redistribution to servers or lists, or reuse of any copyrighted component of this work in other works.}

}

\markboth{Journal of \LaTeX\ Class Files,~Vol.~18, No.~9, September~2020}%
{How to Use the IEEEtran \LaTeX \ Templates}

\maketitle

\begin{abstract}
Manifold learning techniques play a pivotal role in machine learning by revealing lower-dimensional embeddings within high-dimensional data, thus enhancing both the efficiency and interpretability of data analysis by transforming the data into a lower-dimensional representation. However, a notable challenge with current manifold learning methods is their lack of explicit functional mappings, crucial for explainability in many real-world applications. Genetic programming, known for its interpretable functional tree-based models, has emerged as a promising approach to address this challenge. Previous research leveraged multi-objective GP to balance manifold quality against embedding dimensionality, producing functional mappings across a range of embedding sizes. Yet, these mapping trees often became complex, hindering explainability. In response, in this paper, we introduce Genetic Programming for Explainable Manifold Learning (GP-EMaL), a novel approach that directly penalises tree complexity. Our new method is able to maintain high manifold quality while significantly enhancing explainability and also allows customisation of complexity measures, such as symmetry balancing, scaling, and node complexity, catering to diverse application needs. Our experimental analysis demonstrates that GP-EMaL is able to match the performance of the existing approach in most cases, while using simpler, smaller, and more interpretable tree structures. This advancement marks a significant step towards achieving interpretable manifold learning.

\end{abstract}


\section{Introduction}
\IEEEPARstart{M}{any} real-world high-dimensional datasets have highly complex topological structure \cite{Birjandtalab2016,Singh2023,Galelli2013}. Manifold learning (MaL) methods, a category of Nonlinear Dimensionality Reduction (NLDR) techniques, are crucial for transforming these datasets into reduced embedding spaces, thereby aiding in the comprehension of the data's intrinsic structure. Techniques like functional mapping and graph-based methods facilitate the visualization of these embeddings but often result in significant data loss \cite{Gracia2014}. In contrast, non-mapping and deep learning methods, while mitigating data loss, compromise interpretability due to their increased complexity. The application of complex, unexplainable NLDR methods, especially in regulated sectors, can have substantial ethical, legal, scientific, and commercial implications, as outlined in legislation such as the EU's General Data Protection Regulation (GDPR) \cite{GDPR2016a}.

In high-risk domains such as healthcare, it is essential for practitioners to understand the features driving model outcomes \cite{Longo2024,Ali2023,ahmad2018interpretable,Maddigan2022}. 
This necessity underscores the growing importance of interpretable NLDR techniques. Genetic Programming (GP), an evolutionary computation (EC) method, evolves symbolic functional mappings represented by syntax trees. Prior research \cite{GPMaL, GPMaLMO, review} has shown that GP-based methods are effective in creating interpretable manifold learning models, particularly when the symbolic trees maintain low complexity. 
A recent study \cite{Maddigan2024} further extends the explainability of GP trees from manifold learning models by integrating large language models such as ChatGPT to provide conversational explanations. GP, as a mapping method, provides an explicit function enabling the reproduction of embeddings and the implementation of sensitivity analysis. Techniques like automatic program simplification can further simplify these functions. Hence, GP is highly effective in scenarios where understanding the relative importance of features is crucial for ethical and legal reasons.

Interpretability in machine learning, however, remains a subjective concept \cite{Carvalho2019,Singh2024}. Defining exactly what makes a model interpretable is challenging \cite{Molnar2020}, yet there are clear indications of a model being uninterpretable, such as having very large numbers of parameters or complex operations. If we characterise an interpretable model as having a low level of complexity in its tree structure, we can then define complexity by a metric, replacing subjectiveness with a measurable quantity. This, however, creates an inherent trade-off: a less complex model will generally produce lower-quality embeddings. Existing GP-NLDR work \cite{GPMaLMO} used a multi-objective approach to balance embedding quality and the number of embedding dimensions. In this work, we propose an explainability-focused approach, where the complexity of the GP model is explicitly represented as an objective alongside embedding quality.

\subsection*{Major Contributions}
\begin{itemize}
\item We introduce GP-EMaL, a novel GP approach aiming at improving the explainability of GP trees/models for multi-objective manifold learning.
\item Through comprehensive experiments, we demonstrate that GP-EMaL retains performance comparable to existing methods while significantly enhancing interpretability.
\item To support further research and practical applications, we provide open-source code and a web-based application for further experimentation and exploration in this field.
\end{itemize}

\section{Background and Related Work}

\subsection{Dimensionality Reduction}
Dimensionality reduction, a key technique in machine learning, focuses on reducing the number of features in a dataset while preserving as much original information as possible. This process is essential for two main reasons: firstly, it significantly reduces the computational burden associated with large datasets; secondly, it helps overcome the ``curse of dimensionality" \cite{Domingos2012}, a phenomenon where the feature space becomes so sparse in high dimensions that the model's performance degrades, leading to overfitting.

Dimensionality reduction can be classified into two broad approaches: feature selection \cite{10007067} and feature extraction/construction \cite{Zebari2020}. Feature selection involves choosing a subset of relevant features from the dataset. This method selects features without altering their original form, thus maintaining the interpretability of the data.

Feature extraction or construction, on the other hand, transforms the original features into a new set of features. This transformation can be further divided into mapping and non-mapping methods. Mapping methods, such as Principal Component Analysis (PCA), create a new, reduced feature space through an explicit, interpretable functional mapping. This mapping is reversible (and reusable), allowing for a reconstruction of the original features from the reduced set. Non-mapping methods such as t-SNE \cite{van2009dimensionality} and UMAP \cite{mcinnes2020umap}, in contrast, focus on reducing the dimensionality without providing a functional mapping for reversing the process. These methods typically prioritize data compression and efficient representation over interpretability.

\subsection{Nonlinear Dimensionality Reduction}
Nonlinear dimensionality reduction (NLDR) is employed when relationships in a dataset are too complex to be captured by linear methods alone \cite{van2009dimensionality}. Notable among NLDR techniques are deep neural networks, such as autoencoders, which consist of an encoder compressing data into a lower-dimensional space and a decoder reconstructing it back. This process is assessed using a loss function that measures the discrepancy between the input and output, enabling the autoencoder to efficiently map the data while preserving structural integrity \cite{auto}.

Another significant category of NLDR methods is graph-based approaches, such as t-distributed stochastic neighbour embedding (t-SNE) \cite{t-SNE}. These methods create a nearest neighbour graph to capture the inherent structure of the data. The graph is then embedded into a lower-dimensional space, preserving its structural characteristics. t-SNE specifically constructs a graph by considering pairwise similarities between high-dimensional data points and then subsequently optimising a 2-D or 3-D embedding to align with this graph. This method is particularly effective in preserving local data structures and for visualising high-dimensional data \cite{distill}.

The integration of GP in NLDR was pioneered by the GP for Manifold Learning (GP-MaL) method \cite{GPMaL}. Subsequent studies such as GP-MaL-MO \cite{GPMaLMO} have continued to build on the concept, further advancing the research in this field.  GP-Mal-MO applies a multi-objective approach to GP-based NLDR, balancing between preserving the nearest neighbour structure in the embedding and minimising the embedding dimensionality. This paper builds upon these developments to enhance the explainability of GP-based NLDR models.

\subsection{GP and Evolutionary Multi-Objective Optimisation}

GP is an evolutionary computation technique that evolves programs, typically represented as tree structures, to solve problems or model data \cite{9881543}. GP is particularly notable for its ability to evolve interpretable models, making it a valuable tool in domains where understanding the model's decision-making process is crucial \cite{poli2008field,10466603}. In manifold learning, GP has been effectively used to evolve mappings that reduce data dimensionality while preserving its intrinsic structure \cite{GPMaL}.

Evolutionary Multi-Objective Optimisation (EMO) extends the EC paradigm to handle multiple, often conflicting, objectives. EMO algorithms evolve a population of solutions, aiming to find a set of Pareto-optimal solutions that represent the best possible trade-offs between the objectives \cite{coello2006evolutionary}. This approach is particularly beneficial when dealing with complex problems where optimising a single objective could lead to sub-optimal or undesirable solutions.

A critical development in the EMO field was the MOEA/D (Multi-Objective Evolutionary Algorithm based on Decomposition) method \cite{zhang2007moead}. MOEA/D decomposes a multi-objective optimisation problem into a number of scalar optimisation sub-problems and optimises them simultaneously. Each sub-problem focuses on a specific region of the Pareto front, enabling MOEA/D to effectively explore and exploit the search space. This method has gained popularity due to its efficiency and effectiveness, especially in problems with a complex Pareto front landscape.

In the context of GP for manifold learning, MOEA/D offers a robust framework for balancing the trade-offs between manifold quality, embedding complexity, and interpretability. By applying MOEA/D, GP is expected to evolve a diverse set of Pareto-optimal solutions, each representing a different balance between these objectives. This flexibility allows users to select a solution that best fits their specific needs, whether prioritising interpretability for a domain expert's analysis or optimising performance for automated tasks.

\subsection{Tree Complexity}
In GP, the complexity of the syntax trees is a crucial factor affecting both the performance and interpretability of the generated models. One method for calculating tree complexity starts at the root node and progresses recursively \cite{expression}. This method, outlined in \cref{complex_ref}, aggregates the complexity of each node based on specific rules, facilitating efficient ($\mathcal{O}(t)$ for a tree containing $t$ nodes) and consistent calculations (e.g.\ as opposed to time-based complexity measures \cite{time}). While this approach generates simpler trees, it has its limitations: it does not constrain the shape of the embedding tree, potentially resulting in asymmetrical and visually confusing/uninterpretable structures. Asymmetrical trees are harder to understand than the equivalent symmetrical tree with the same number of nodes, as they contain more layers of nested functions. Additionally, while this existing approach penalises complex operations (function nodes), it does not explicitly minimise the tree size, often leading to large trees using many simpler operators. The fixed function set also does not consider the varying interpretability of functions across different contexts.

\begin{table}[tb]
\vspace{-1em}
 \caption{Complexity Values \cite{expression}}
 \label{complex_ref}
\begin{center}
\begin{tabular}{ ll} 
\hline \noalign{\smallskip}
Complexity($n$) & Symbol of node$(n) $ \vspace{2pt}\\
\hline \noalign{\smallskip}
1 & constant\vspace{2pt}\\
 2 & variable\vspace{2pt}\\
  $\sum_{c \in c_n}\text{Complexity}(c)$  & $+, -$ \vspace{2pt}\\
   $\prod_{c \in c_{n}}\text{Complexity}(c)$ + 1  & $*, /$ \vspace{2pt} \\
Complexity$(n_1)^2$ &  square \vspace{2pt}\\
Complexity$(n_1)^3$  & square root \vspace{2pt}\\
$2^{\text{Complexity}(n_1)}$& sin, cos, tan, exp, log 
\\\noalign{\smallskip}

 \hline
\end{tabular}
\end{center}
where $c$ is a child of node $n$, and $n_{1}$ is the first child of $n$. 
\vspace{-1em}
\end{table}

\subsection{Tree Complexity Metrics}
Various tree complexity metrics have been proposed in the literature, falling into two main categories: structural complexity and functional (semantic) complexity metrics \cite{Le2016,chadalawada, campobello, Tang}. Structural complexity assesses the tree at a node level, while functional complexity is based on the complexity of the tree's overall behaviour.

Research on structural complexity has explored factors like penalizing large trees through parsimony pressure \cite{Soule1998EffectsOC} and reducing nested functions \cite{Smits2005ParetoFrontEI}. However, the integration of other metrics, such as assessing the asymmetry of trees, has been less explored. Our work aims to fill this gap.

In terms of functional complexity, the prevalence of many nonlinear operators, especially when nested, has often been identified as a factor of complexity \cite{expression}. The use of Tikhonov regularization to apply a global smoothing function to the tree has also been studied \cite{tik}. This method penalizes the function based on the norm of its higher-order derivatives, resulting in a smoother function.

The distinction between structural and functional complexity is not absolute, as these aspects can overlap. For example, a tree represented by a nested function like $sin(cos(exp(x)))$ may be considered both functionally complex (due to its nonlinear nature) and structurally complex (due to its nested structure). Thus, these complexity measures should be viewed as complementary heuristics rather than distinct categories. For measuring explainability, structural complexity approaches are generally more appropriate as they directly consider the complexity of the tree structure, which is what the user will be trying to interpret.

\section{Proposed Method: GP-EMaL}
\begin{figure*}[!t]
  \vspace{-2em}
  \hspace{-1.3em}\includegraphics[width=1.05\textwidth]{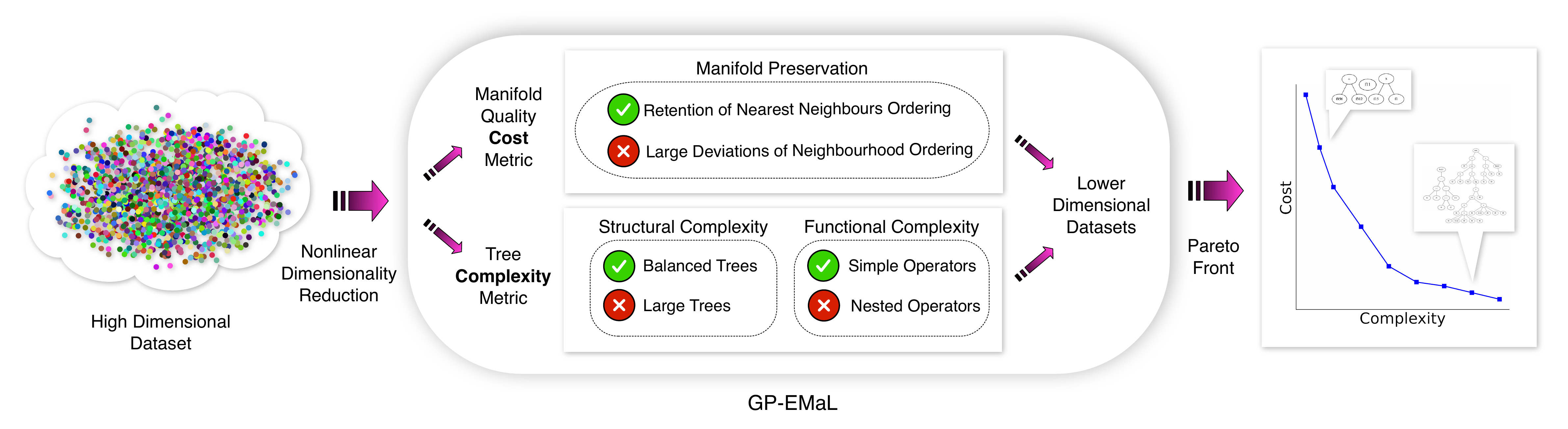}
  \caption{GP-EMaL Architecture}
  \label{overview}
\end{figure*}
In our proposed method, GP-EMaL, we build upon the framework established by GP-MaL-MO \cite{GPMaLMO}, targeting a key limitation: the tendency towards complex tree structures which can impede interpretability. Our GP-EMaL approach introduces a novel complexity metric for expression trees, replacing the manifold dimensionality objective in GP-MaL-MO that focuses on minimizing the complexity while retaining the essential manifold quality cost metric based on neighbourhood ordering \cite{GPMaLMO,GPMaL}. This change aims to enhance the tree interpretability without compromising the effectiveness of the manifold learning process. The core algorithm for multi-objective optimisation is similar to GP-MaL-MO, as illustrated in \cref{overview}, which outlines the architecture of GP-EMaL. 

\subsection{GP Design}

The evolutionary process begins by randomly initialising a population of GP individuals. These individuals are evaluated according to two objectives: one measuring manifold embedding quality and the other measuring tree complexity. The MOEA/D algorithm is used to find the population Front, which is used to select individuals for breeding in the next generation. This process repeats for a number of generations, with the Front produced by the final generation representing the different trade-offs between manifold quality and GP individual complexity.

GP-EMaL, like GP-MaL-MO, uses a multi-tree structure where each individual contains several trees. Each of these trees produces a single dimension of the lower-dimensional embedding. An individual is randomly initialised with a number of trees in the range $[2, m \div 2]$ for $m$ features. By utilising a specialised mutation operator (``Add/Remove Tree"), individuals are able to add and remove trees during evolution. This should allow the evolutionary search to balance manifold embedding quality with the number of trees and tree complexity. We also retain a ```Standard" mutation operator, which picks a random subtree from any tree in the GP individual to perform mutation on. Likewise, crossover picks a random subtree from each of the two individuals being mated.

\subsection{Complexity Metric}
\label{metric_section}
The complexity metric in GP-EMaL has been enhanced to include parameterisation of the function set, allowing users to tweak the complexity cost of operators according to their specific needs/domain. For instance, linear operations like addition and subtraction could be assigned lower costs due to their straightforward interpretation in financial contexts, such as calculating balances or changes in value. Conversely, in theoretical physics applications, complex mathematical functions like trigonometric, exponential, and logarithmic operators are frequently used to model and understand phenomena such as wave mechanics, quantum physics, or signal processing. These complex operators, though less intuitive for a general audience, are essential and interpretable within the context of these specialized fields, as they accurately represent the underlying principles and behaviours being studied. We also provide a set of default parameters 

To further improve interpretability, our new metric also focuses on achieving balanced tree structures through an asymmetry penalty term ($A$) and encourages the development of smaller trees via a step-wise scaling term ($S$). These improvements in GP-EMaL capture several crucial dimensions of tree complexity:
\begin{enumerate}
    \item     \textbf{Structure}: The $A_i$ term encourages tree structures to be balanced, improving interpretability.
    \item     \textbf{Size}: Smaller trees, being easier to interpret, are favoured. The $S_i$ scaling term proportionally increases the cost for larger trees.
    \item     \textbf{Content}: Our metric allows for a strategic selection of operators within the tree, prioritising those that enhance simplicity and context-specific understandability.
    \item     \textbf{Semantics}: The placement of complex operators within the tree directly impacts its interpretability. Placing complex operators closer to the leaves of the tree improves interpretability by avoiding nested complex functions. By using a hierarchical cost assignment to operators, our method aims to avoid complexity near the leaves and simplify the overall structure.

\end{enumerate}

This refined metric in GP-EMaL enhances the interpretability of the evolved trees, addressing critical aspects of tree complexity while preserving the efficacy of the manifold learning process. We discuss the design of each part of this metric in the following subsections.

\subsubsection{Symmetry Balancing Term} 
\label{structure_section}
In GP-EMaL, the symmetry balancing term $A_{i}$ is applied to each node, calculated based on the size disparity between the left and right subtrees, denoted as $\text{Size}_\text{Left}$ and $\text{Size}_\text{Right}$ respectively. This term is crucial for promoting symmetrical tree structures, as it imposes a penalty proportional to the asymmetry between subtrees. So that symmetric trees are not penalised, a deduction of 1 is applied to the term. The formula for this penalty is:


\begin{equation}
A_{i} = 2^{|\Delta_i|} - 1
\label{penality_eq}
\end{equation}

\noindent where  $\Delta = \text{Size}_\text{Left} - \text{Size}_\text{Right}$. This method contrasts with GP-MaL-MO, where trees often become unbalanced and visually complex. The application of $A_{i}$ at every node ensures that balance is considered at all levels of the tree, leading to structures that are easier to interpret and analyse.

\subsubsection{Scaling term} 
 Recognising that larger trees pose a challenge to interpretability, GP-EMaL incorporates a scaling term $S_{T}$ to penalise excessive tree sizes. The term is governed by the parameter $\alpha$, representing tree T's size ($t$) relative to the pre-defined maximum tree size (i.e.\ $\text{Size}_{\text{Max}}$). The punishment for larger trees begins when $\alpha$ exceeds a predefined threshold $\mu$, set by default to 0.75. An increase in $\mu$ results in a stronger penalty for larger trees. The scaling term is defined as:

\begin{equation}
\quad {S_{T}} = 
\begin{cases}
1, & \text{if } \alpha_{i} < \mu,\\
2\alpha_{i}, & \text{if } \alpha_{i} > \mu
\end{cases}
\quad
\alpha_{i} = \frac{t}{\text{Size}_{\text{Max}}}
\quad
\label{S_eqn}
\end{equation}

\begin{figure}[!t]
  \vspace{-1em}
  \centering
  \includegraphics[width=0.45\textwidth]{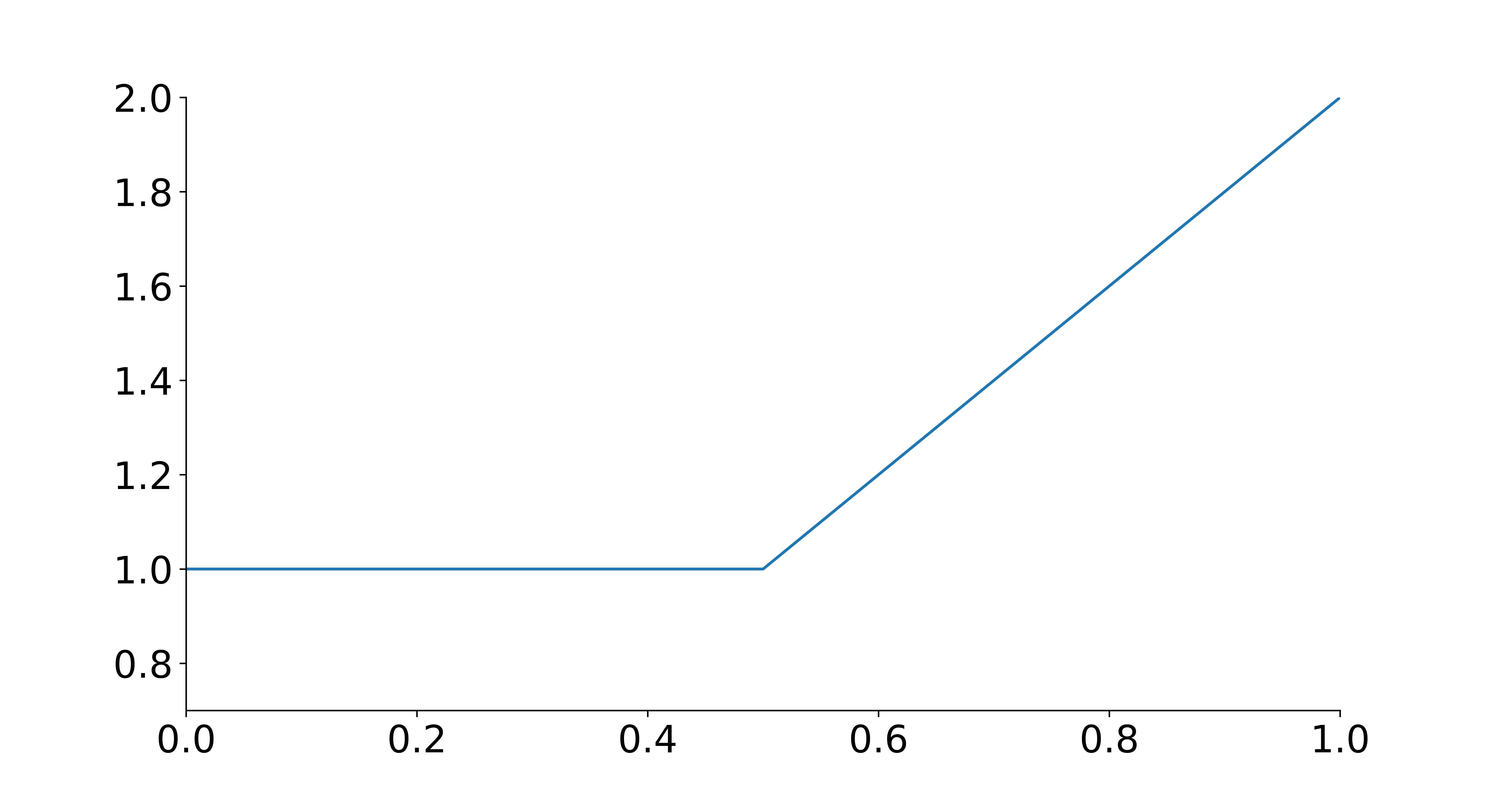}
  \caption{Scaling term for S([0,1]) with $\mu=0.5$}
  \label{size_plot}
  \vspace{-1em}
\end{figure}

\subsubsection{Tree Complexity Term} 
The GP-EMaL approach introduces enhanced flexibility in measuring tree complexity by introducing a parameterisable function set. This set allows users to define both the tree operators and their respective contributions to overall tree complexity. The cost assigned to each operator is proportional to the height of the respective subtree, with the scaling of these costs categorised as linear, quadratic, or exponential.

For example, consider a function set comprising three basic arithmetic operators and a nonlinear operator: $[+,-,\times, sigmoid]$, paired with a function cost set $[sum, sum, prod, exp]$. This configuration results in a higher prevalence of the simpler arithmetic operators $[+,-]$ within the evolved trees due to their lower cost. The multiplication operator $\times$ would appear less frequently, typically higher in the tree, while the nonlinear operator $sigmoid$ would be least common, generally positioned near the top of the tree, directly affecting the features.

This methodological approach in GP-EMaL allows for nuanced control over the complexity of the generated trees. It enables the specification of not only which operators are present but also their preferred locations within the tree structure, thereby influencing both the functional complexity and the interpretability of the trees.

\begin{table}[!t]
 \caption{Summary of Function Cost Sets}
\begin{center}
\begin{tabular}{ cccc } 
 \hline \noalign{\smallskip}
Cost & Operator & Complexity & Cost Scaling \vspace{2pt}\\
 \hline \noalign{\smallskip}
sum & $n_{i} \in (+,-)$  &$L_{i} + R_{i}$ & $\mathcal{O}(t)$ \\
prod & $n_{i} \in (\times,\div)$ & max($L_{i}\times R_{i},L_{i},R_{i}$) & $\mathcal{O}(t^{2})$  \\
exp & $n_{i} \in (f_{1},..,f_{n})$ & $2^{(L_{i} + R_{i})}$ & $\mathcal{O}(2^{t})$ \\\noalign{\smallskip}
 \hline
\end{tabular}
\end{center}
\label{function_cost_sets}
  \vspace{-1em}
\end{table}

\Cref{function_cost_sets} summarises the settings for these function cost sets. Here, the complexity of the left and right subtrees is denoted as $L_{i}$ and $R_{i}$, respectively, with $f_{1},..,f_{n}$ representing the set of non-arithmetic or ``special" operations (e.g.\ trigonometric functions). The notation $\mathcal{O}$ indicates the asymptotic scaling complexity in relation to the average size (number of nodes, $t$) of the left and right child subtrees based on a binary tree structure.

The tree complexity function, denoted as $F(T)$, is defined as:

\begin{equation}
\begin{split}
    F(T) &= 
    S_{T}\times[\sum_{n_{i} \in (+,-)} (L_{i} + R_{i} + A_{i})\\
    &+ \sum_{n_{j} \in (\times,\div)} (\text{max}(L_{j}\times
    R_{j}, L_{j}, R_{j}) + A_{j})\\
    &+ \sum_{n_{k} \in (f_{1},..,f_{n})} (2^{(L_{k} + R_{k})} + A_{k})]
\end{split}
\label{complexity_eq}
\end{equation}

\noindent where $A_{i}$ refers to the asymmetry penalty at each node as defined in \cref{penality_eq}, and $S_{t}$ is the scaling term for tree size as per \cref{S_eqn}.

\subsection{Function and terminal sets}
Our default function set contains a range of basic and more complex operators. It includes the arithmetic functions ($+$,$-$,$\times$, protected $\div$), conditional functions (max and min), and nonlinear operators (absolute value, ReLU and the Sigmoid function).  The terminal set (leaf nodes) contains each of the features of the dataset.

\subsection{Objective functions}
As a multi-objective GP-MaL approach, our method has two objectives: one that minimises the difference between the input feature space and the embedding space; and a second that minimises the complexity of the trees used to transform the input space into the embedding space.

As our focus in this work is on reducing the complexity of the learned trees, we utilise the same first objective (Cost) as in GP-MaL-MO, which has shown to be an effective measure:

\begin{equation}
    Cost(I,X) = \frac{1}{|X|} \sum_{x \in X} \frac{(1 - \text{corr}(N,N'))}{2}
\label{gpmalmo_eq}
\end{equation}

for GP individual $I$ and dataset $X$ with data point $x\in X$. corr represents the Spearman's rank correlation coefficient, $|X|$ is the number of data points, and the $2$ in the denominator is a scaling factor. This equation penalises GP individuals where the embedding has a different neighbourhood structure from the input dataset. 

Whereas GP-MaL-MO simply uses embedding dimensionality (number of trees) as the complexity measure, our proposed GP-EMaL uses the new complexity metric introduced in \cref{metric_section}. Given that a GP-EMaL individual contains multiple trees (embedding dimensions), we sum across the tree complexity metric (\cref{complexity_eq}) to get the overall complexity of a given GP individual ($I$):

\begin{equation}
    Complexity(I) = \sum_{T \in I}F(T)
\label{L_eqn}
\end{equation}

\noindent where $T$ is a tree in $I$.

These two objective functions are generally conflicting, as less complex trees represent functions with fewer degrees of freedom and, therefore, cannot retain the neighbourhood structure of the input space as accurately as more complex trees, leading to a higher cost. By using the MOEA/D algorithm to evolve GP individuals according to these two objectives, we are able to produce a population of GP individuals to obtain an approximate Pareto front, where individuals represent different (non-dominated) tradeoffs between embedding quality (Cost) and tree complexity.

\subsection{Implementation}
\begin{figure}[tbp]
  \centering
  \includegraphics[width=0.495\textwidth]{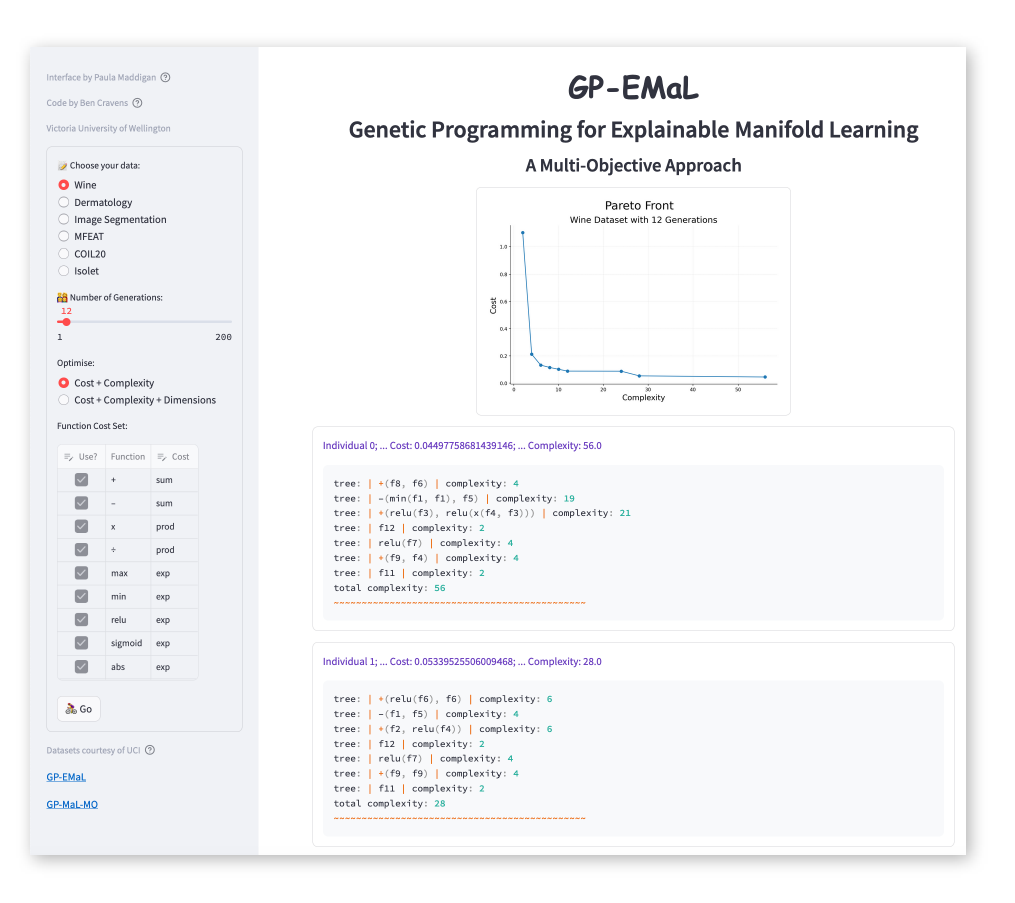}
  \caption{Streamlit Application Interface.}
  \label{streamlit}
\end{figure}

The GP-EMaL algorithm is built using Python with the code publicly available\footnote{https://github.com/cravies/GP-EMaL}. A Streamlit web-based application is also accessible for ease of use\footnote{https://gp-emal.streamlit.app/} depicted in \cref{streamlit}.

\section{Experiment Design}
We evaluated our new GP-EMaL approach using the following methodology, split into three different analyses. 

\subsection{Predictive Performance}
In our study, we adopt a widely used approach for comparing manifold learning (MaL) algorithms, focusing on the classification accuracy attainable in the embedded spaces they produce. This method avoids biases associated with unsupervised comparisons based on specific cost or quality formulations. While other supervised learning tasks like regression could also serve as indicators of MaL quality, classification is particularly effective in evaluating the preservation of data structures in embeddings. 

Our evaluation involves training KNN and RF classifiers on embeddings from both the proposed GP-EMaL and baseline GP-MaL-MO methods, using ten-fold cross-validation with random forest (RF) and k-nearest neighbour (KNN) classifier. The random forest, comprising an ensemble of 100 trees, is chosen for its reliable and stable performance, making it a suitable tool for our analysis. KNN was selected as a simple method that heavily relies on high-quality embeddings for achieving high classification accuracy, due to its distance-based approach.

This evaluation utilized seven datasets, listed in \cref{datasets}, primarily from the UCI repository \cite{UCI}, and included the COIL20 dataset \cite{Nene1996}. These datasets were chosen as a broadly representative set containing varying numbers of features, instances, and classes. Classification accuracy was recorded over the 30 runs for each dataset with parameter settings as outlined in \cref{parameters}. 

\subsection{Complexity Summary Statistics} We also gathered summary statistics across all the datasets studied. For both GP-EMaL and GP-MaL-MO, we measured the average number of nodes, the number of unique features used, and the complexity of the nodes chosen. This provides a high-level comparison of the complexity/explainability of the two methods.

\subsection{Visual Comparisons} Finally, to more directly contrast the explainability of the two approaches, we compare a typical example individual generated by each of GP-MaL-MO and GP-EMaL on the COIL-20 dataset. This dataset is the largest one (by number of features) in our tests and so represents a very difficult challenge for producing an explainable NLDR model.  

\begin{table}[!t]
 \caption{Classification Datasets for Evaluation}
\begin{center}
\begin{tabular}{ lrrr } 
\hline \noalign{\smallskip}
Dataset & Features & Instances & Classes \vspace{2pt}\\
\hline \noalign{\smallskip}
Wine &13& 178 &3 \\
Dermatology &34 &358 &6 \\
Image Segmentation &19& 2310 &7\\
MFEAT &649& 2000 &10\\
MNIST & 784 &2000& 2 \\
COIL20 &1024 &1440& 20\\
Isolet&617&1560&26  \\\noalign{\smallskip}

 \hline 
\end{tabular}
\end{center}
\label{datasets}
\end{table}

\begin{table}[!t]
 \caption{GP-EMaL Parameter Settings}
\begin{center}
\begin{tabular}{ ll l ll} 
\hline \noalign{\smallskip}
Parameter & Setting \vspace{2pt}\\
\hline \noalign{\smallskip}
Generations & 1000 \\
Population Size & 100 \\
“Standard” Mutation & 15\% \\
Add/Remove Tree Mutation &15\% \\
Crossover  & 70\% \\
Max. No. Trees& max(2, $[m\div2]$)\\
Min. Tree Depth &2 \\
Max. Tree Depth& 14\\
Decomposition &Tchebycheff\\
Pop. Initialisation &Ramped Half-and-half\\
Function Set & $+ - \times \div$ 
            ReLu sigmoid max min abs\\
Cost Set & sum sum prod prod
         exp exp exp exp exp \\\noalign{\smallskip}
 \hline
\end{tabular}
\end{center}
\label{parameters}
\end{table}

\section{Results}

\subsection{Predictive Performance}
We begin by exploring the trade-off between complexity and classification accuracy in the individuals produced by GP-EMaL in \cref{partOne}. We do not directly compare with GP-MaL-MO on our proposed complexity metric, as it would represent an unfair comparison --- we would expect GP-MaL-MO to perform substantially worse given that, unlike GP-EMaL, it was not optimised according to this metric. 

Instead, we compare and contrast the performance of GP-EMaL and GP-MaL-MO according to simple measures of tree complexity in \cref{partTwo}. These include the number of nodes, the number of complex operators (those with exponential cost scaling), and the number of unique features used by the GP individuals. While none of these provides a wholly unbiased measure of complexity, analysing these different measures provides a clearer view of the differences between our proposed GP-EMaL method and the GP-MaL-MO baseline.
\subsubsection{GP-EMaL Complexity vs.\ Classification Accuracy}
\label{partOne}

\begin{figure}[!t]
  \centering
  \vspace{-1em}
\includegraphics[width=0.495\textwidth]{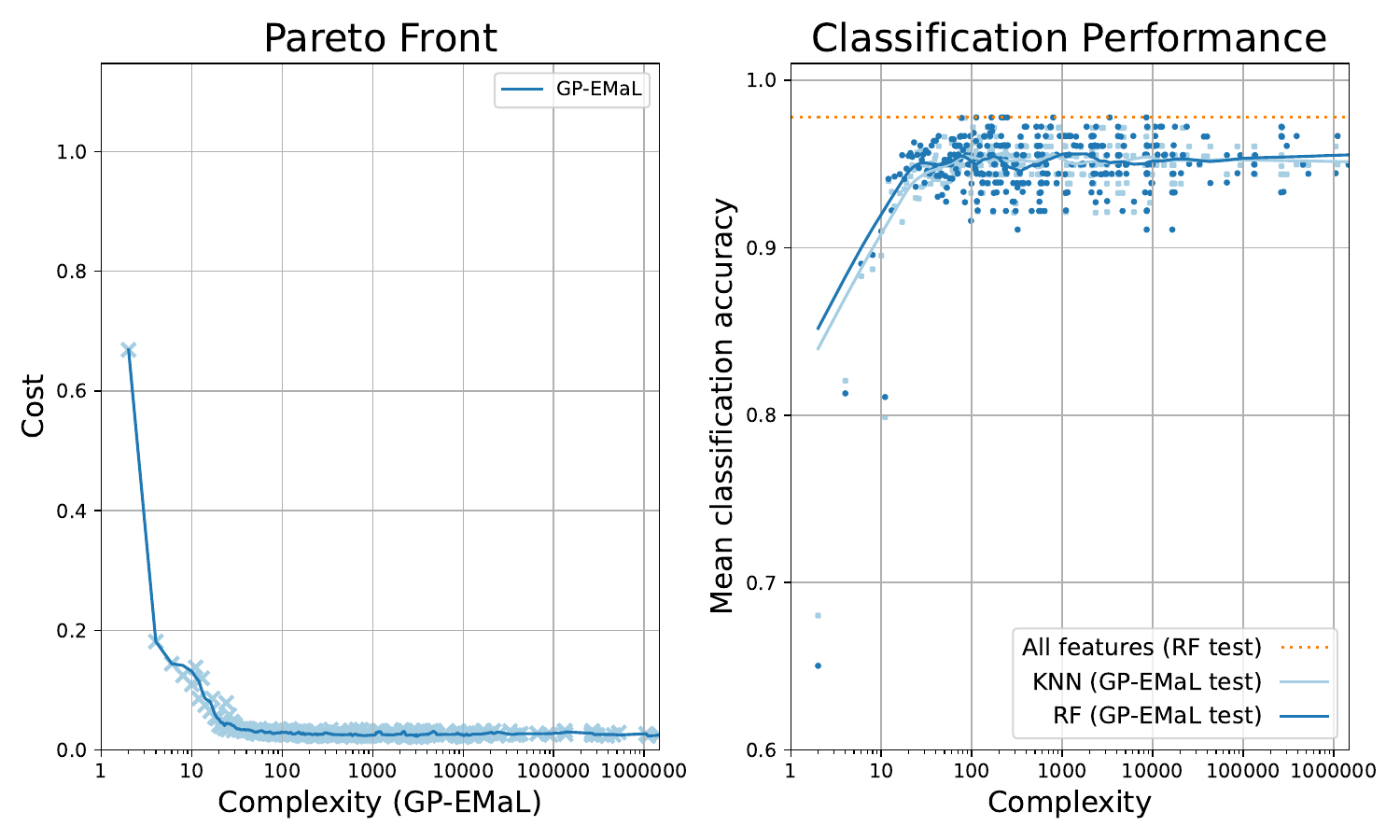}
        \vspace{-1em}
  \caption{Front and Classif.\ Performance on {Wine}}
  \label{wine_front}
\end{figure}

\begin{figure}[!t]
  \centering
 \vspace{-1em} \includegraphics[width=0.495\textwidth]{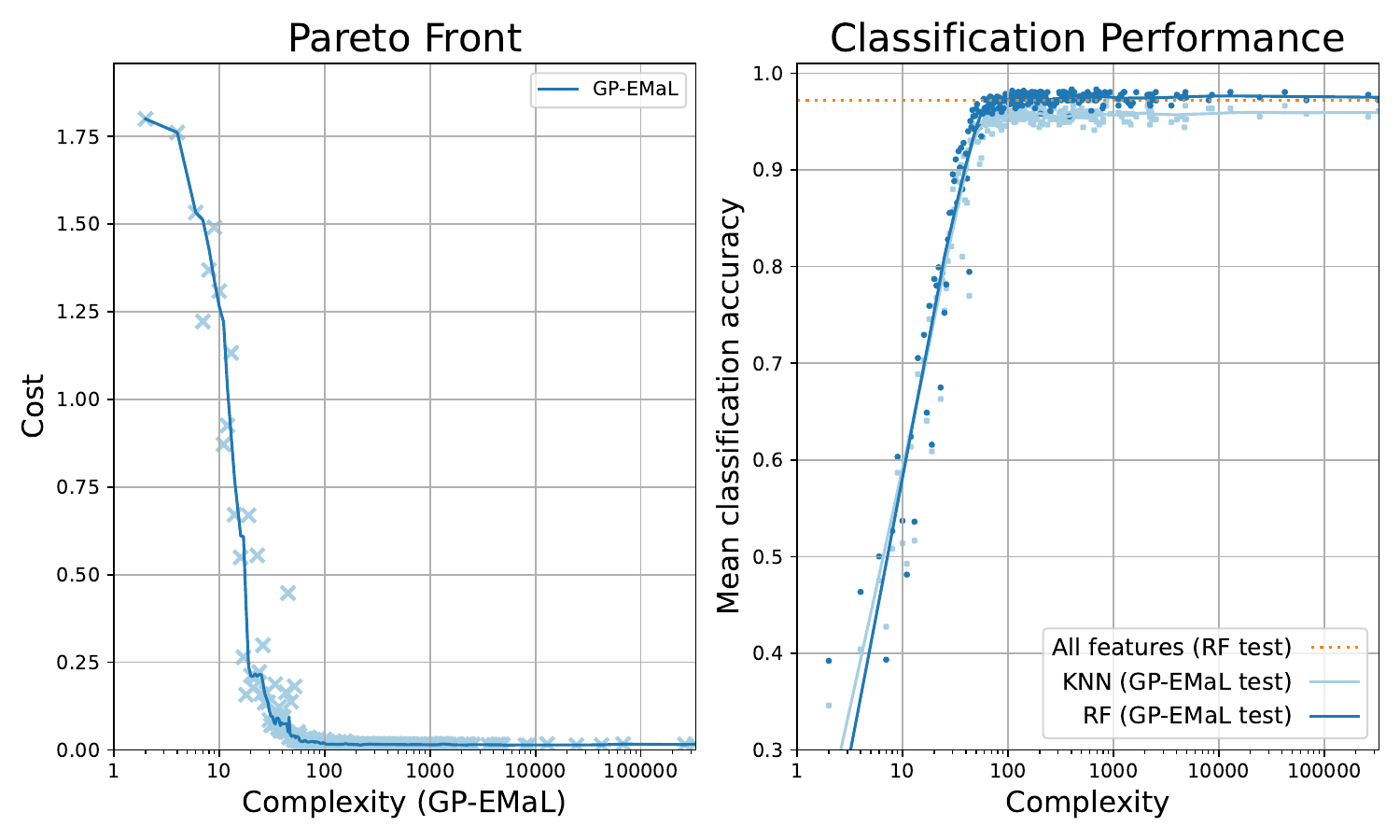}
      \vspace{-1em}
  \caption{Front and Classif.\ Performance on {Dermatology} }
  \label{derm_front}
\end{figure}

\begin{figure}[!t]
  \centering
    \vspace{-1em}
\includegraphics[width=0.495\textwidth]{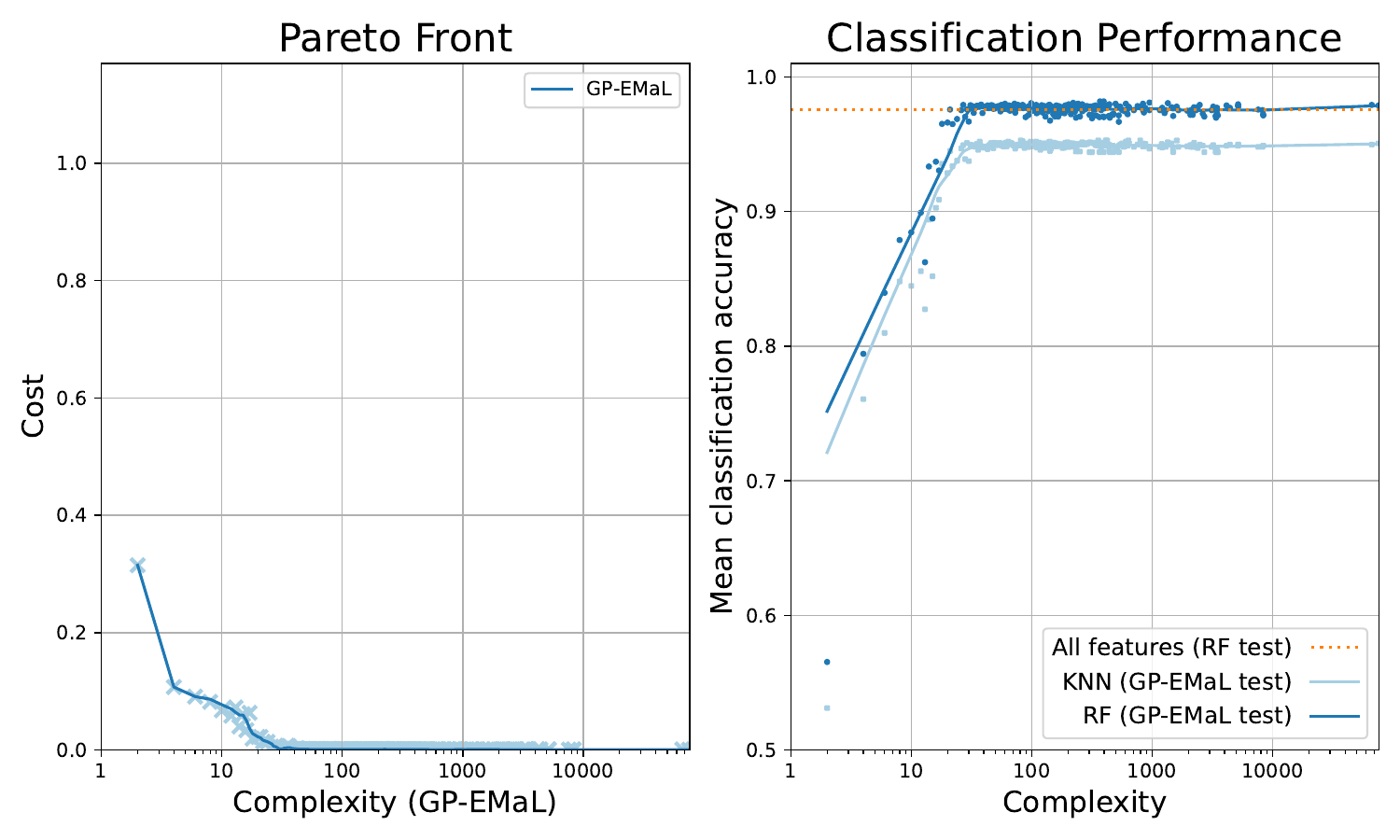}
        \vspace{-1em}
  \caption{Front and Classif.\ Performance on {Image Segmentation} }
  \label{image_front}
\end{figure}

\begin{figure}[!t]
  \centering
          \vspace{-1em}
\includegraphics[width=0.495\textwidth]{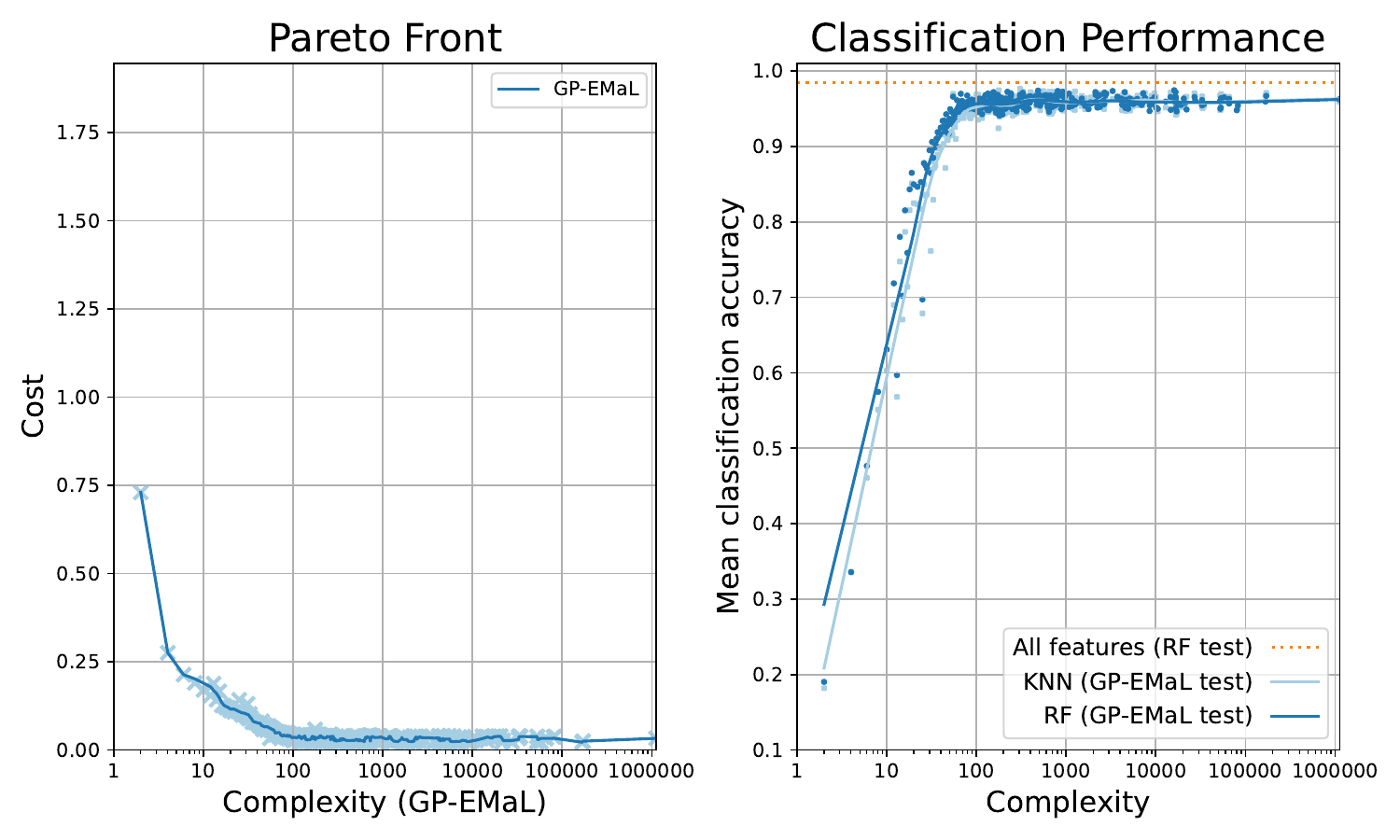}
        \vspace{-1em}
  \caption{Front and Classif.\ Performance on the {MFEAT} dataset}
  \label{mfat_front}
\end{figure}

\begin{figure}[!t]
  \centering
  \vspace{-1em}
\includegraphics[width=0.495\textwidth]{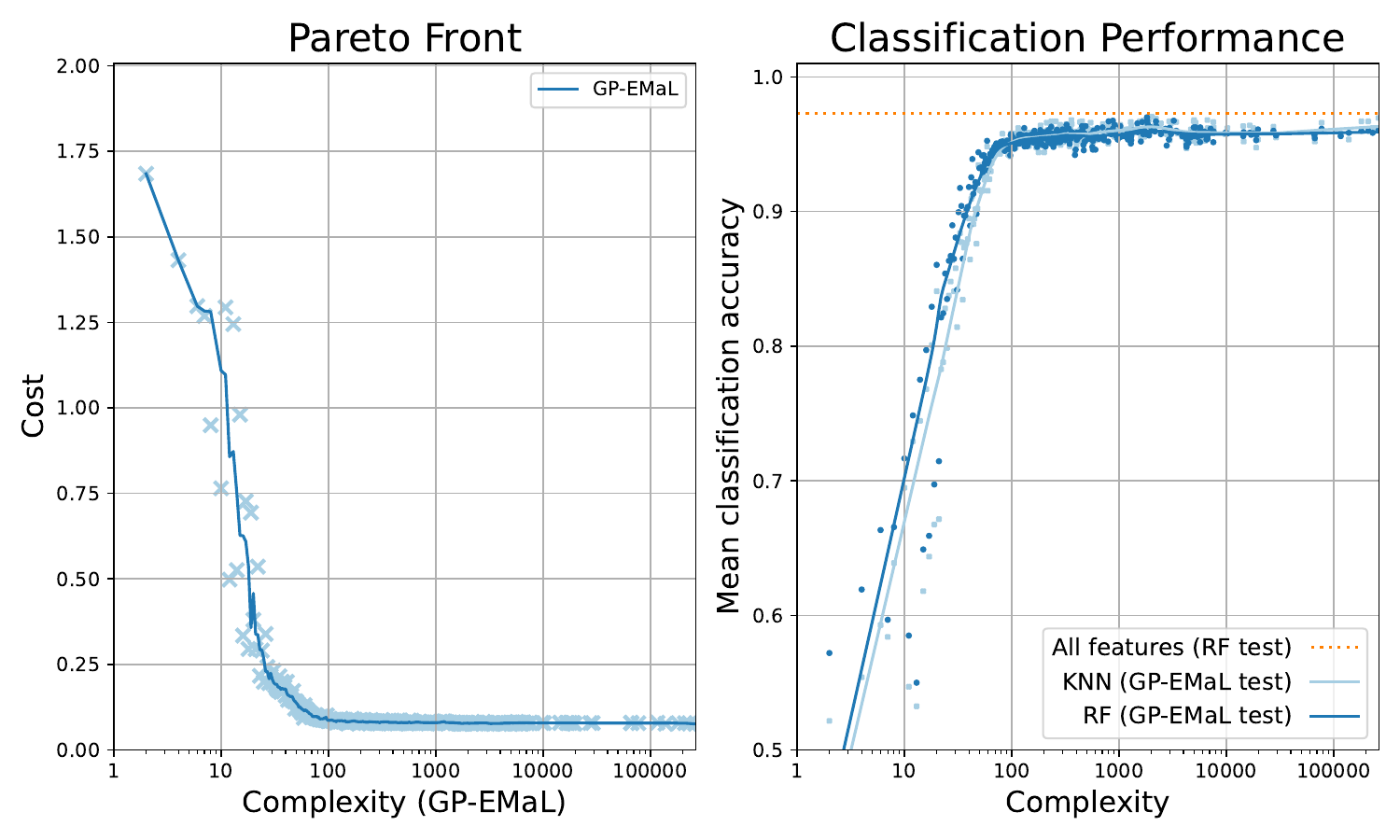}
        \vspace{-1em}
  \caption{Front and Classif.\ Performance on {MNIST}}
  \label{mnist_front}
\end{figure}

\begin{figure}[!t]
  \vspace{-1em}
  \centering
\includegraphics[width=0.495\textwidth]{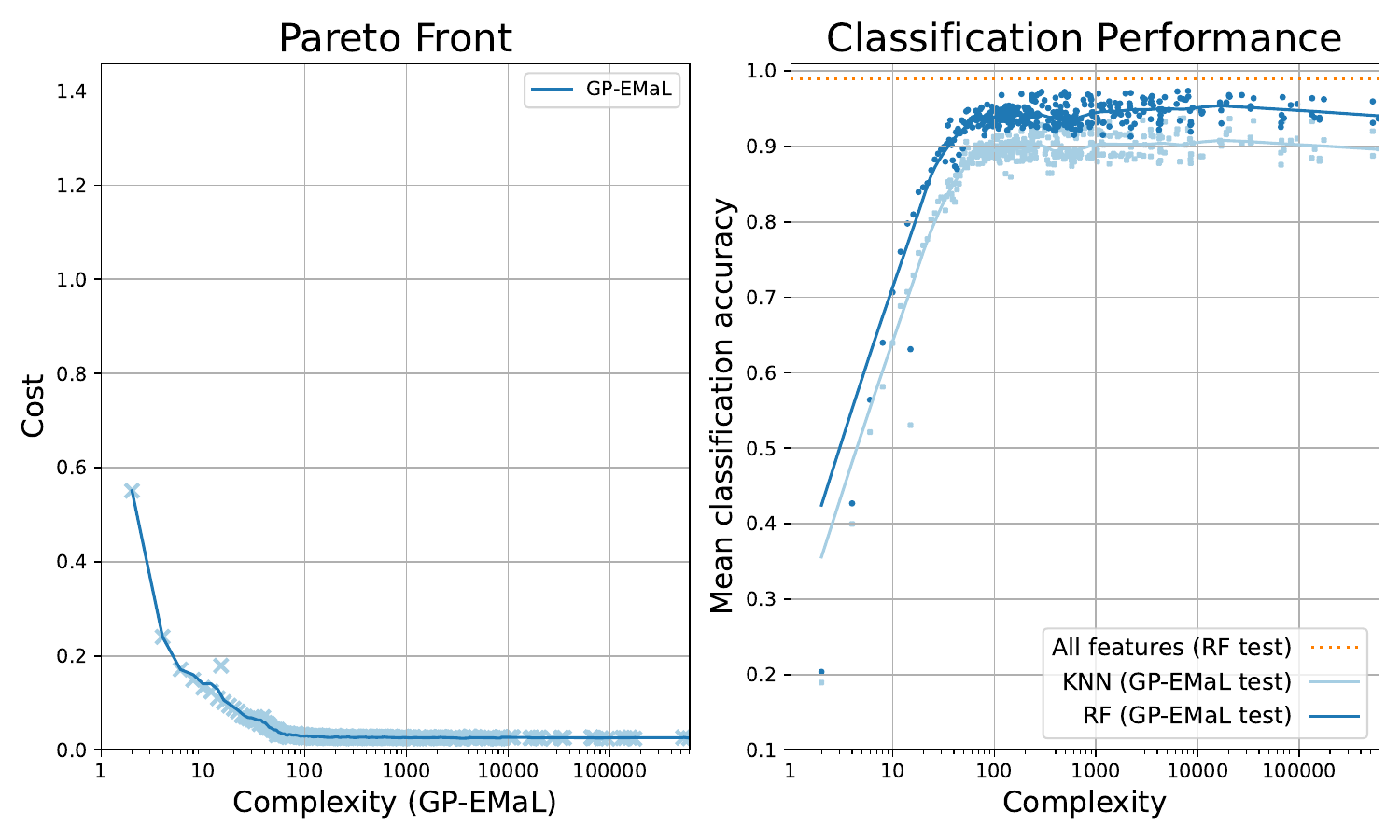}
       \vspace{-1em}
  \caption{Front and Classif.\ Performance on {COIL20} }
  \label{coil_front}
\end{figure}

\begin{figure}[!t]
  \vspace{-1em}
  \centering
\includegraphics[width=0.495\textwidth]{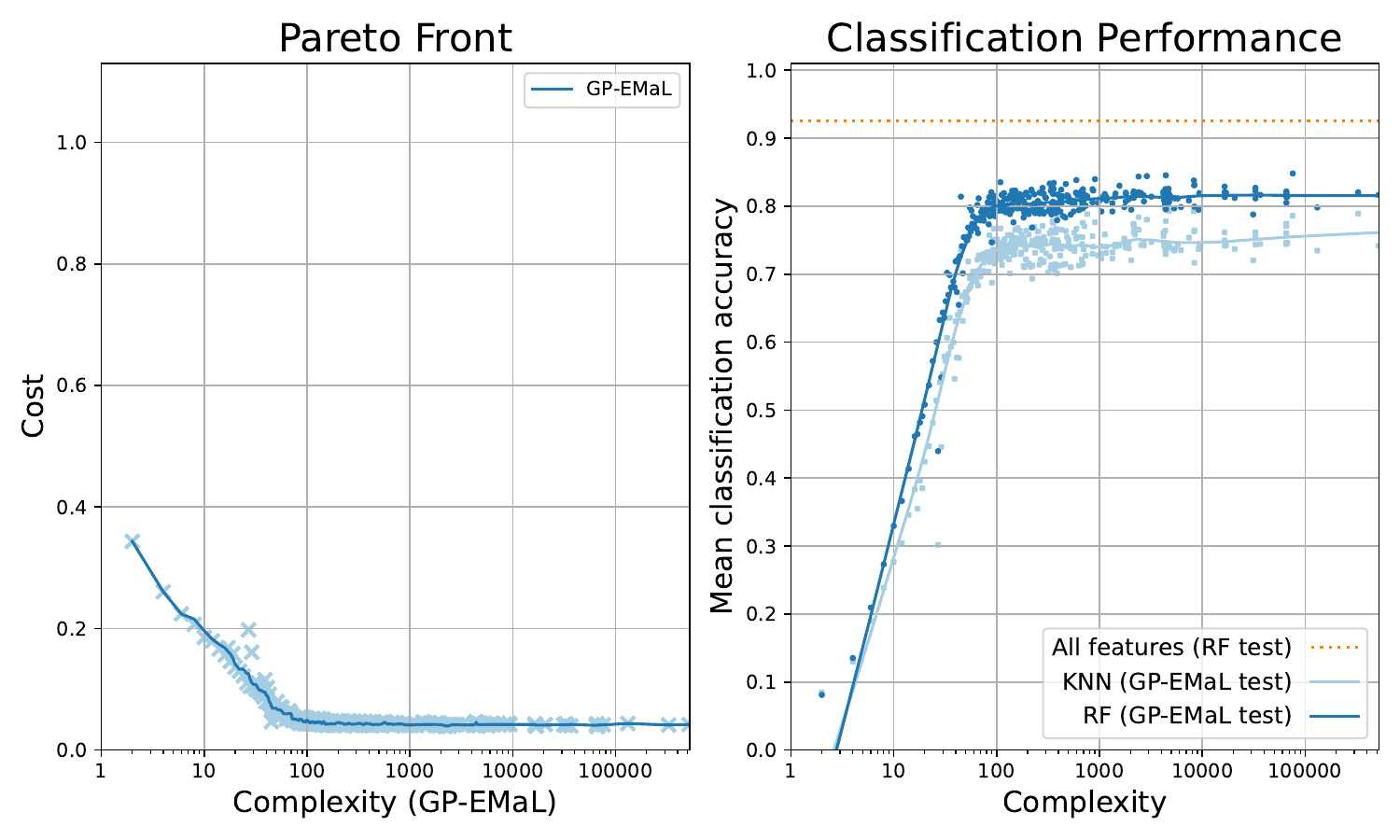}
        \vspace{-1em}
  \caption{Front and Classif.\ Performance on {Isolet}}
  \label{isolet_front}
\end{figure}

\Crefrange{wine_front}{isolet_front} show the results for GP-EMaL using the seven datasets, in order of roughly increasing dataset complexity. For each dataset, we show two visualisations. The first depicts the approximated Pareto front. The cost (reduction in manifold quality) is measured on the y-axis, with the tree complexity metric (in log scale) shown on the x-axis. As both cost and complexity should be minimised, points closer to the bottom-left are of higher quality. The second plot shows the Classif.\ Performances of both KNN (lighter blue) and RF (darker blue) predictors using the new embeddings. A smoothing spline is run on the results to average and smooth the curves over the 30 runs. As classification accuracy should be maximised, points closer to the top-left of the second set of plots are better-performing. We also include a dotted orange horizontal line, which is the performance of the RF classifier when using all features. This line represents a benchmark of what a good classifier could achieve on the source high-dimensional dataset.


The Wine dataset (\cref{wine_front}) is the least complex of our tested datasets, with only 13 features and 178 instances. The approximated Pareto front shows that GP individuals with very low complexity (${\approx}$30 as measured by \cref{gpmalmo_eq} ) are able to achieve a very low cost, strongly retaining the structure of the original high-dimensional space by using only simple operators. A similar result is seen in the classification plots, where both the KNN and Random Forest classifiers perform well with above 90\% accuracy using the embeddings produced by a simple GP individual.

The Dermatology dataset (\cref{derm_front}) exhibits similar behaviour to that of Wine 
but requires a slightly more complex solution (${\approx}$75) to generate an embedding with a cost near zero. Both RF and KNN achieve in excess of 90\% classification accuracy, matching the baseline performance of all features.

The Image Segmentation results (\cref{image_front}) resemble those of Wine, given the similar number of features in both datasets. This is the first set of results where there is a substantial performance difference between RF and KNN. The RF classifer achieves over 95\% accuracy with a GP complexity of only ${\approx}$30, whereas the KNN classifier is slightly worse across all complexities. 

The MFEAT and MNIST datasets show comparable patterns in their results (\crefrange{mfat_front}{mnist_front}). They show convergence to low-cost solutions at a slightly higher complexity ($\approx$100) than the simpler datasets, achieving over 90\% classification accuracy on RF and KNN. These are the first datasets where there is a (very small) gap in performance between RF using all features and RF using the embedding produced by even the most complex GP trees. This is an inherent trade-off: on sufficiently complex problems, a more complex (less explainable) model is needed to achieve the best performance. However, as we show in the rest of this paper, a much simpler model often can still achieve very high performance --- a trade-off that is easily preferable in many applications. 

The most complex datasets, COIL20 and Isolet (\crefrange{coil_front}{isolet_front}) also show an asymptotic pattern, where cost approaches zero at a complexity of ${\approx}75$. While the RF classification performance on COIL20 is near the baseline, on Isolet, there is a gap of around $~10\%$, even at the highest complexities. If very high performance is crucial on Isolet, the original GP-MaL-MO approach may be required (as discussed later) instead. 

\subsubsection{Comparing GP-EMaL to GP-MaL-MO}
\label{partTwo}

To compare GP-EMaL to GP-MaL-MO as fairly as possible, we evaluate the performance of each method along three different signals of complexity/interpretability. These are: 
\begin{enumerate}
    \item The total number of nodes used in the GP individual, which gives a ``raw" measure of tree complexity;
    \item The total number of \textit{exp} cost operators used in the GP individual, which measures how many of the most difficult-to-understand functions are used; and 
    \item The number of unique features used, i.e.\ the number of dimensions in the high-dimensional input data that were used to construct the embedding. Using fewer unique features will make a GP individual easier to understand, as it requires less interpretation of different aspects of the data domain.  
\end{enumerate}

\begin{figure*}[tbp]
  \centering
  \includegraphics[width=\textwidth]{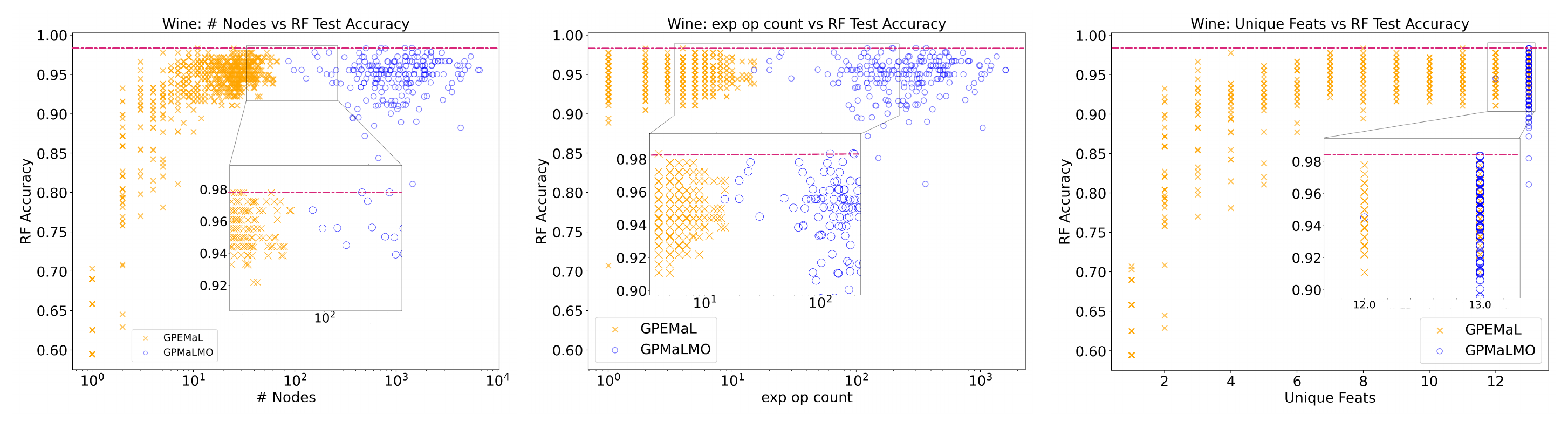}
   \caption{Performance Comparisons using the {Wine} dataset}
  \label{wine_plots}
\end{figure*}

\begin{figure*}[tbp]
  \centering
  \includegraphics[width=\textwidth]{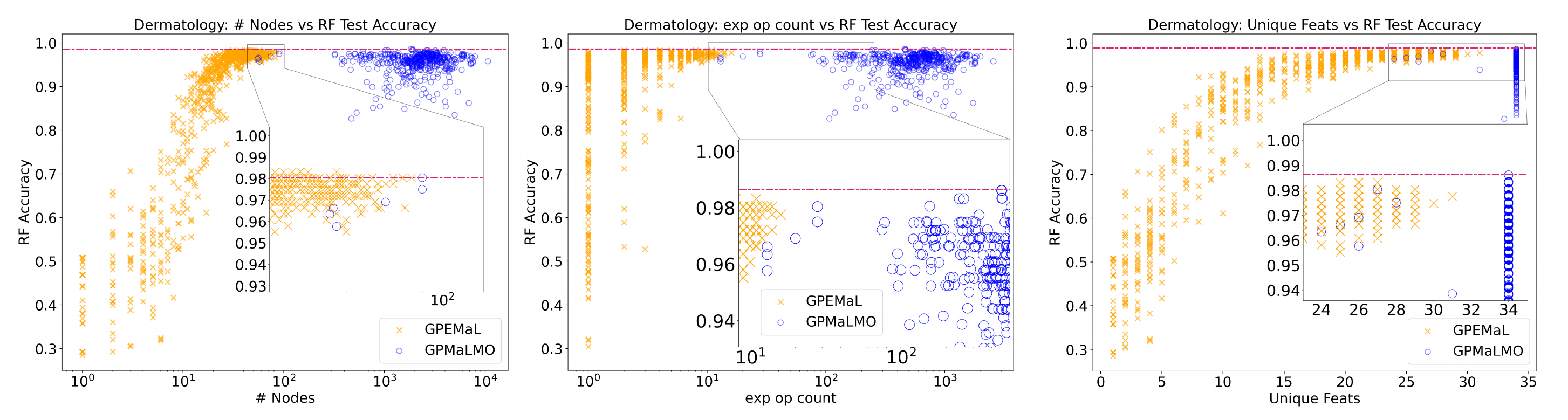}
  \caption{Performance Comparisons using the {Dermatology} dataset}
  \label{derm_plots}
\end{figure*}

\begin{figure*}[tbp]
  \centering
  \includegraphics[width=\textwidth]{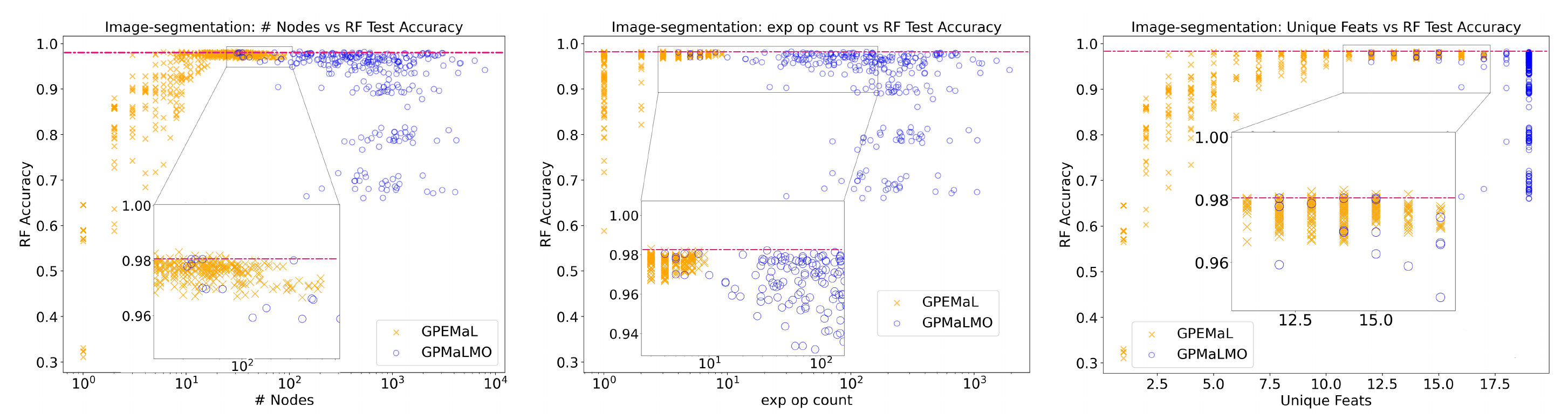}
   \caption{Performance Comparisons using the {Image Segmentation} dataset}
  \label{image_plots}
\end{figure*}


\begin{figure*}[tbp]
  \centering
  \includegraphics[width=\textwidth]{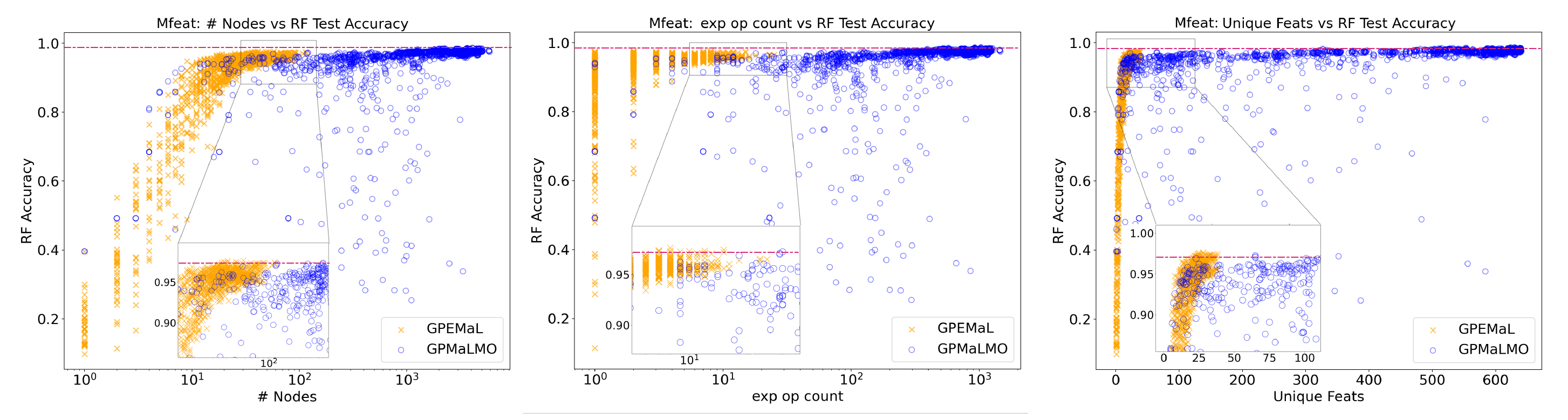}
   \caption{Performance Comparisons using the {MFEAT} dataset}
  \label{mfat_plots}
\end{figure*}

\begin{figure*}[tbp]
  \centering
  \includegraphics[width=\textwidth]{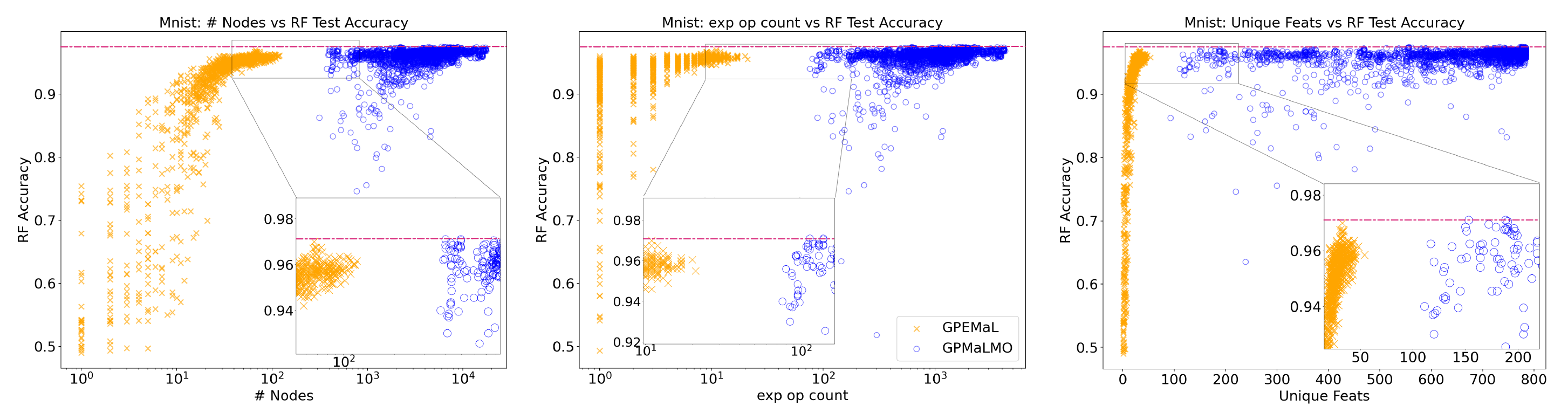}
   \caption{Performance Comparisons using the {MNIST} dataset}
  \label{mnist_plots}
\end{figure*}

\begin{figure*}[tbp]
  \centering
  \includegraphics[width=\textwidth]{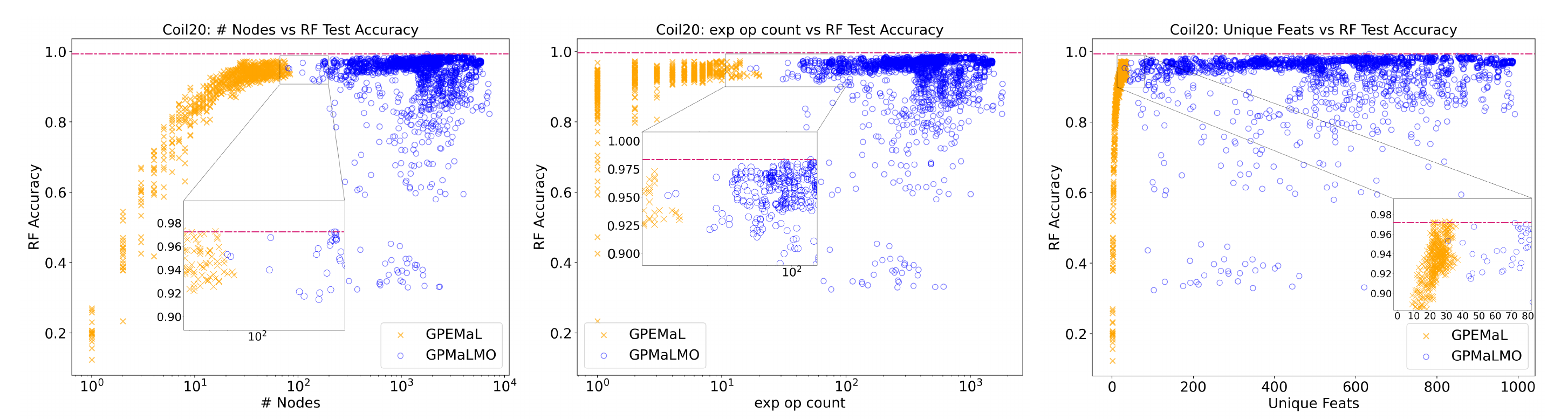}
  \caption{Performance Comparisons using the \textit{COIL20} dataset}
  \label{coil20_plots}
\end{figure*}

\begin{figure*}[tbp]
  \centering
  \includegraphics[width=\textwidth]{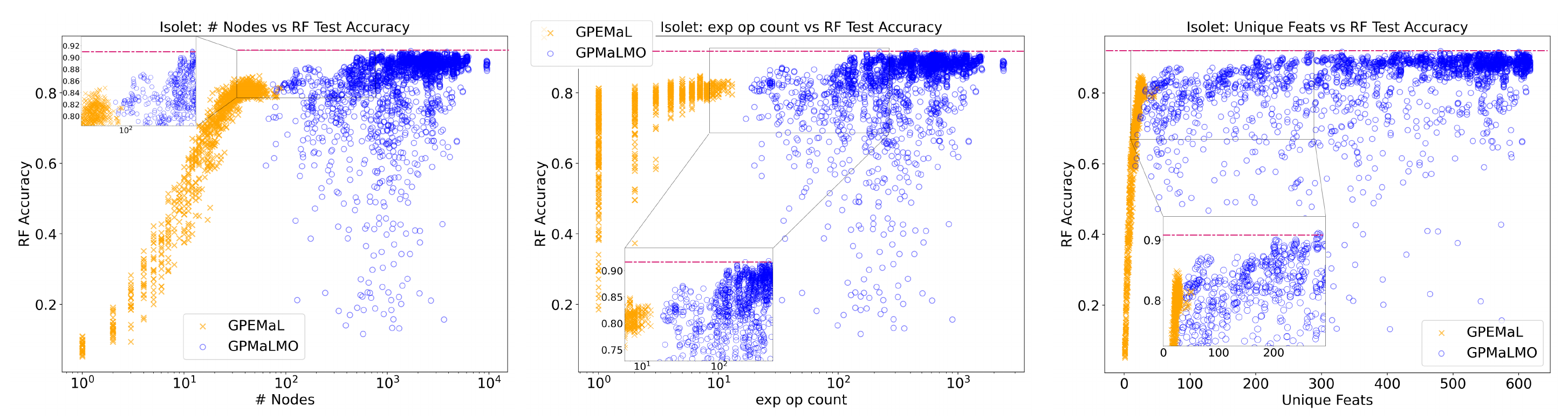}
  \caption{Performance Comparisons using the {Isolet} dataset}
  \label{isolet_plots}
\end{figure*}

We plot each of these measures against the classification accuracy for all individuals in the fronts evolved by both GP-EMaL and GP-MaL-MO across all our runs. We use only RF for computing classification accuracy, as it always outperforms KNN and hence provides a more accurate measure of embedding quality. The plots for each of the datasets are shown in \crefrange{wine_plots}{isolet_plots}, where each orange {\color{orange}$\bm{\times}$} symbol shows one GP individual evolved by GP-EMaL and each {\color{blue}$\bm{\circ}$} symbol shows one GP individual evolved by GP-MaL-MO. We also include an overlay within each plot to show a closer perspective of the boundary between the GP-EMaL and GP-MaL-MO regions. For all of these plots, points closer to the top-left of the plot are of higher quality (having higher classification accuracy at a lower complexity). As with our earlier results, we include a dotted horizontal line to represent the ``baseline" performance, which uses the same RF classifier with all the original features of the dataset.

Across nearly all of the datasets, the proposed GP-EMaL method is able to achieve similar performance to the existing GP-MaL-MO method despite having smaller trees with fewer complex operators and using fewer unique features. This is a clear testament to the complexity function proposed in this work --- by using a more sophisticated measure of complexity (compared to just the number of trees in GP-MaL-MO) as our second objective, we can produce explainable trees that effectively preserve data structure in the embedded space.

This improvement in GP-EMaL is especially clear-cut in the first three datasets (\crefrange{wine_plots}{image_plots}). On these, GP-EMaL is able to retain sufficient manifold structure to reach the baseline accuracy while clearly using much less complex trees than GP-MaL-MO --- in other words, GP-EMaL effectively dominates GP-MaL-MO across the three different complexity measures. GP-EMaL is consistently less complex at high levels of accuracy; on the Image Segmentation dataset, for example, GP-MaL-MO has accuracies as low as 70\%, despite using all features in the dataset and using trees with many more nodes. Finally, GP-EMaL provides a much better approximation of the low-complexity part of the Pareto front: it produces extremely simple individuals (e.g.\ using only a few nodes and/or one or two features) that often can achieve over 80\% accuracy --- a trade-off that may be desirable on some problem domains.

On the MFEAT dataset (\cref{mfat_plots}), we observe that GP-MaL-MO has a small but consistent decrease in RF accuracy for less-complex GP individuals. At similar levels of complexity, GP-EMaL outperforms GP-MaL-MO. For example, GP-MaL-MO achieves around 93\% accuracy for a GP individual with 100 nodes, whereas GP-EMaL achieves close to the baseline accuracy (around 97\%) with as few as 40 nodes. Similar patterns are seen for the number of exponential operators and unique features. Indeed, GP-EMaL can consistently reach the baseline accuracy when using only 25 unique features, whereas GP-MaL-MO has significant variance in accuracy.

As the complexity of our datasets further increases, a small gap in classification accuracy between the two methods begins to emerge. On MNIST (\cref{mnist_plots}), GP-EMaL is very slightly ($\leq 1\%$) below the baseline accuracy, whereas GP-MaL-MO is able to reach baseline performance. There is, however, a marked complexity gap between the two methods: GP-EMaL is nearly an order of magnitude less complex than GP-MaL-MO (e.g.\ $100$ vs $500$ nodes) despite the only 1\% difference in performance. A very similar result is observed on the COIL20 dataset (\cref{coil20_plots}), albeit with a smaller gap in complexity and performance between the most-complex GP-EMaL individual and the least-complex GP-MaL-MO individual.

On the most complex dataset, Isolet (\cref{isolet_plots}), GP-MaL-MO is clearly able to attain higher classification accuracy than GP-EMaL --- but not without using more complex GP trees. GP-MaL-MO requires very complex trees (${\approx}1000$ nodes, ${\approx}100$ complex operators, ${\approx}300$ unique features) to achieve a baseline level of accuracy. It is interesting to note that together, GP-EMaL and GP-MaL-MO almost form a combined front, with neither method having individuals that Pareto-dominate the other. On large, complex datasets such as Isolet, it may be desirable to run both GP-EMaL and GP-MaL-MO and produce a combined front to provide the user with a comprehensive set of trade-offs from very simple to very complex models. 

In conclusion, the results demonstrate a clear trade-off between complexity and predictive accuracy across various datasets. This trade-off is more evident in datasets with higher complexity. For most datasets, GP-EMaL offers an improvement over GP-MaL-MO by producing substantially more explainable GP individuals without compromising on predictive accuracy. For the more complex datasets, GP-MaL-MO may have an edge in accuracy, but GP-EMaL stands out as a robust alternative, especially when the focus is on interpretability, with only a minimal performance decrease.

\begin{table}[!tb]
\caption{Summary Statistics}
\begin{center}
\begin{tabular}{p{2cm}p{5.8cm}}
\hline \noalign{\smallskip}
Statistic & Description \vspace{2pt} \\ 
\hline \noalign{\smallskip}
\# $\mathcal{O}(2^{n})$ nodes & total number of $\mathcal{O}(2^{n})$ cost operators \\  
\# $\mathcal{O}(n^{2})$ nodes & total number of $\mathcal{O}(n^{2})$ cost operators \\
\# $\mathcal{O}(n)$ nodes & total number of $\mathcal{O}(n)$ cost operators \\
\# $\mathcal{O}(1)$ nodes & total number of $\mathcal{O}(1)$ cost operators (feature nodes) \\
\# nodes & total number of nodes \\
\# unique feat & number of unique feature nodes \\\noalign{\smallskip}
\hline
\end{tabular}
\label{stats_table}
\end{center}
\vspace{-1em}
\end{table}

\begin{figure}[t!]
    \vspace{-1em}
    \centering
   \hspace{-1em} \includegraphics[width=.495\textwidth]{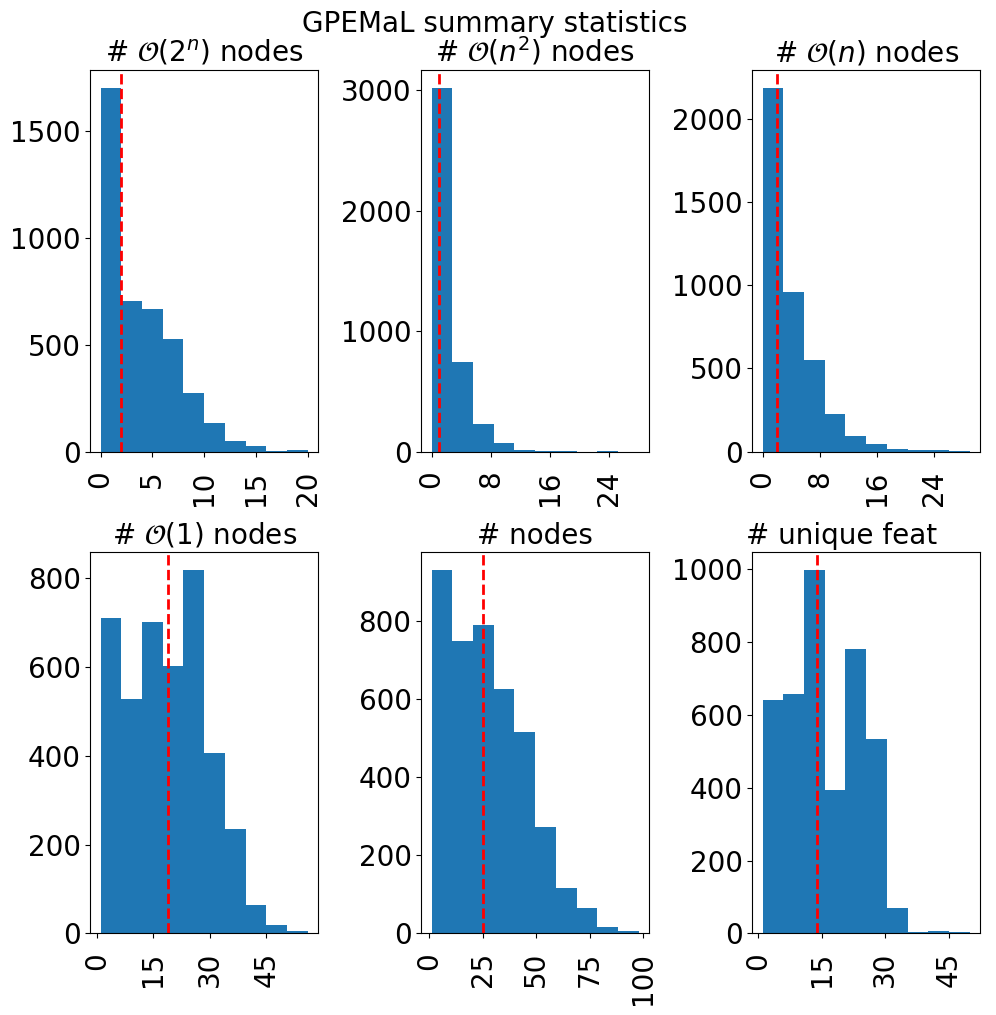}\\ 
   \vspace{1em}
  \hspace{-1em} \includegraphics[width=.499\textwidth]{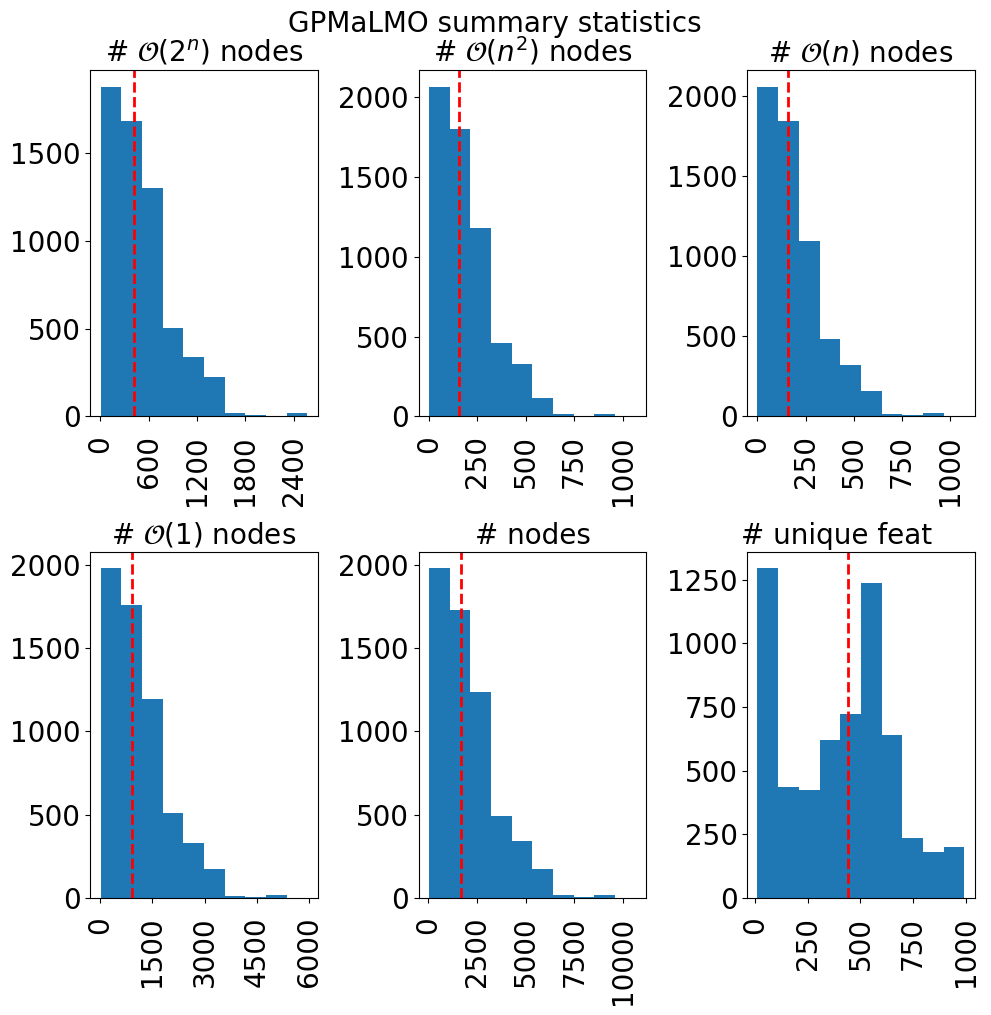} %
  \caption{Summary statistics for the two GP methods across all datasets. The red line represents the median result.}
  \label{stats}
  \vspace{-1em}
\end{figure}

\subsection{Summary Statistics}
To understand the overall complexity of the two methods across all the datasets, we calculated several summary statistics for both GP-MaL-MO and GP-EMaL. These statistics are described in \cref{stats_table}.

\cref{stats} shows the summary statistics in the form of histograms for each of the two GP methods. By looking at the x-axis values, we can see that GP-MaL-MO produces exceptionally more complex trees than GP-EMaL. Indeed, the maximum value for GP-EMaL on each of the graphs is actually below the \textit{median} value on the corresponding graph for GP-MaL-MO. The median and maximum number of nodes used by GP-EMaL are around $100\times$ smaller than that of GP-MaL-MO.   

\subsection{Visual Comparisons}

\begin{figure}[!t]
  \centering
  \includegraphics[width=.48\textwidth]{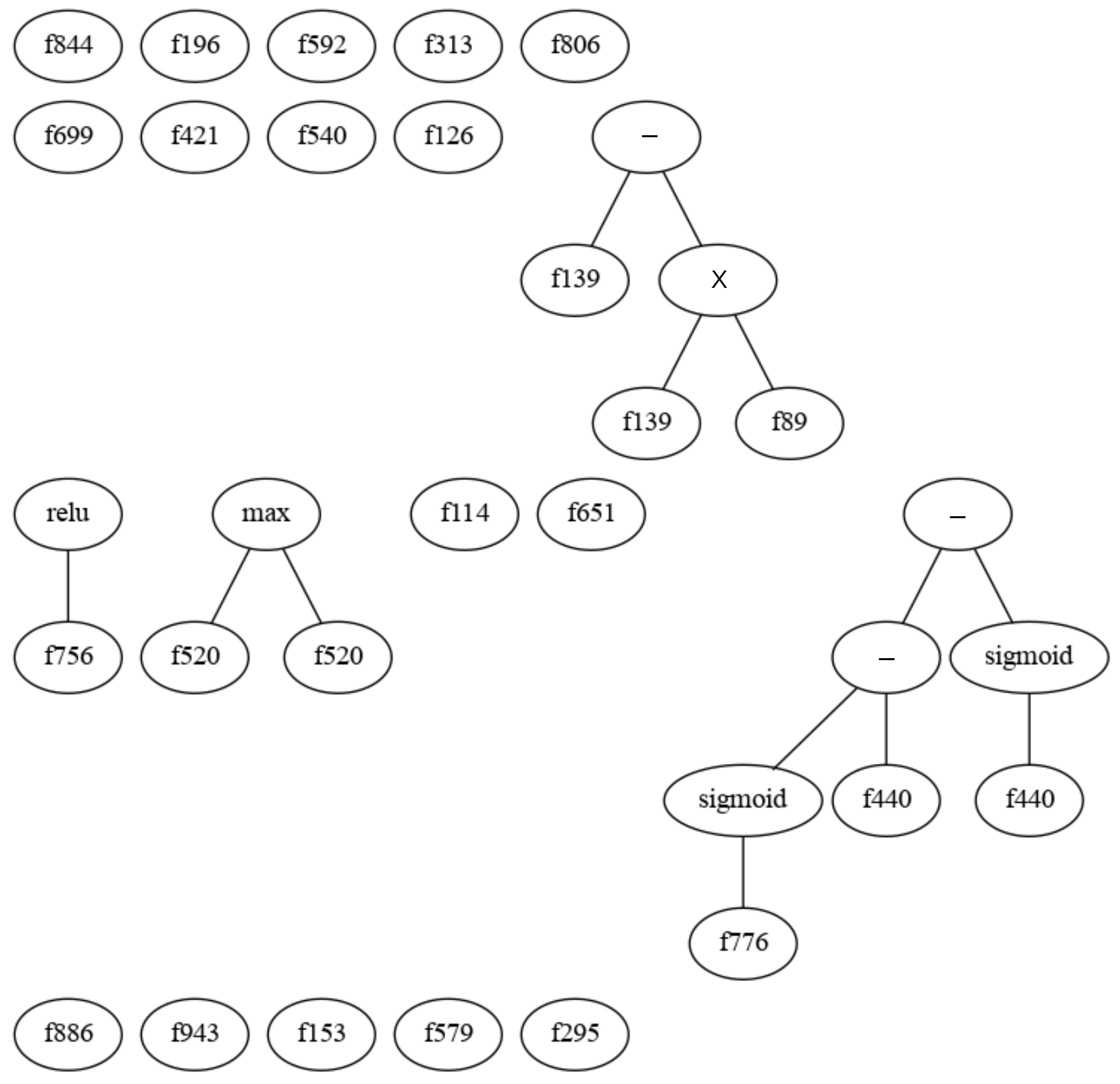}
  \caption{A typical GP-EMaL individual containing 20 trees on COIL20, with 94.2\% test accuracy and a complexity of 78.}
  \label{eg_gpemal}
\vspace{1em}
   \centering
   \includegraphics[width=.53\textwidth]{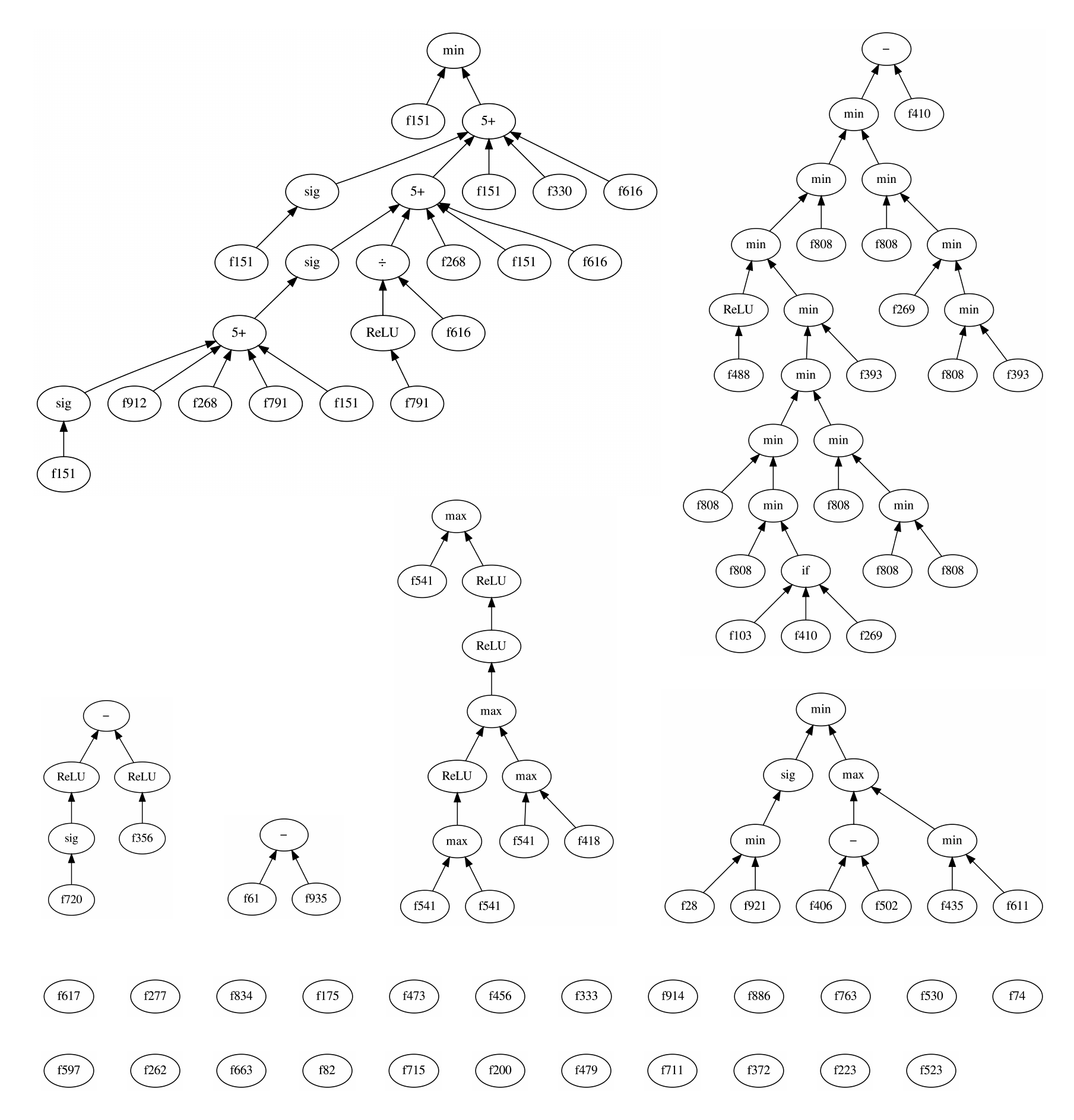}
   \caption{A typical GP-MaL-MO individual containing 29 trees on COIL20, with 99.1\% test accuracy.}
   \label{eg_gpmalmo}
  \vspace{-1em}
\end{figure}

Finally, we present a typical individual evolved by GP-EMaL (\cref{eg_gpemal}) and GP-MaL-MO (\cref{eg_gpmalmo}) on the complex COIL20 dataset\footnote{Further sample tree outputs are available at github.com/cravies/GP-EMaL}. We chose the individual with median classification accuracy across all runs to represent the ``elbow" of the approximated Pareto Front. While GP-EMaL performs slightly worse than GP-MaL-MO on classification accuracy (mirroring our earlier results), the individual itself is markedly less complex: the biggest tree produced by GP-EMaL has seven nodes, compared to 31 in GP-MaL-MO. The deepest tree in GP-EMaL is only a depth of four; GP-MaL-MO produces a tree with a depth of ten. While GP-MaL-MO may be technically \textit{interpretable} to a domain expert, it is clearly very difficult to \textit{explain} compared to the one produced by GP-EMaL. We can also see that the tree structure in GP-EMaL is much more interpretable: the most complex sigmoid and ReLU operators are always at the bottom of the tree, with more human-understandable subtraction and multiplication operators used nearer the root.

\section{Conclusions}

This study's exploration into GP-EMaL represents a significant stride in the realm of interpretable manifold learning using genetic programming. GP-EMaL distinguishes itself by its emphasis on interpretability, a critical aspect often overlooked in traditional approaches. Notably, while GP-EMaL demonstrates a slight decrease in performance on higher-complexity datasets, this reduction is generally modest, typically under 5\%. This is a small price to pay considering the complexity of previous methods, which were so intricate that they were impractical for tasks where interpretability is paramount.

Furthermore, our findings underline the delicate balance between interpretability and performance in machine learning models. The open-source availability of GP-EMaL is intended to foster more research and practical applications in this field. Looking ahead, future work could involve user testing to validate interpretability and investigate ways to enhance embedding quality, potentially through the development of more efficient and specific evolutionary search operators. Furthermore, expanding GP-EMaL to optimise three objectives --- neighbourhood structure preservation, embedding complexity, and dimensionality --- simultaneously would offer a more nuanced approach, allowing for solutions tailored to specific requirements and contexts.

\bibliographystyle{IEEEtran}
\IEEEtriggeratref{36}
\bibliography{refs}

\begin{thebibliography}{10}
\providecommand{\url}[1]{#1}
\csname url@samestyle\endcsname
\providecommand{\newblock}{\relax}
\providecommand{\bibinfo}[2]{#2}
\providecommand{\BIBentrySTDinterwordspacing}{\spaceskip=0pt\relax}
\providecommand{\BIBentryALTinterwordstretchfactor}{4}
\providecommand{\BIBentryALTinterwordspacing}{\spaceskip=\fontdimen2\font plus
\BIBentryALTinterwordstretchfactor\fontdimen3\font minus \fontdimen4\font\relax}
\providecommand{\BIBforeignlanguage}[2]{{%
\expandafter\ifx\csname l@#1\endcsname\relax
\typeout{** WARNING: IEEEtran.bst: No hyphenation pattern has been}%
\typeout{** loaded for the language `#1'. Using the pattern for}%
\typeout{** the default language instead.}%
\else
\language=\csname l@#1\endcsname
\fi
#2}}
\providecommand{\BIBdecl}{\relax}
\BIBdecl

\bibitem{Birjandtalab2016}
J.~Birjandtalab, M.~B. Pouyan, and M.~Nourani, ``Nonlinear dimension reduction for eeg-based epileptic seizure detection,'' in \emph{2016 IEEE-EMBS International Conference on Biomedical and Health Informatics (BHI)}, 2016, pp. 595--598.

\bibitem{Singh2023}
D.~Singh, H.~Climente-Gonzalez, M.~Petrovich, E.~Kawakami, and M.~Yamada, ``Fsnet: Feature selection network on high-dimensional biological data,'' in \emph{2023 International Joint Conference on Neural Networks (IJCNN)}, 2023, pp. 1--9.

\bibitem{Galelli2013}
S.~Galelli and A.~Castelletti, ``Tree-based iterative input variable selection for hydrological modeling,'' \emph{Water Resources Research}, vol.~49, no.~7, pp. 4295--4310, 2013.

\bibitem{Gracia2014}
A.~Gracia, S.~González, V.~Robles, and E.~Menasalvas, ``A methodology to compare dimensionality reduction algorithms in terms of loss of quality,'' \emph{Information Sciences}, vol. 270, pp. 1--27, 2014.

\bibitem{GDPR2016a}
\BIBentryALTinterwordspacing
{European Parliament} and {Council of the European Union}. Regulation ({EU}) 2016/679 of the {European} {Parliament} and of the {Council}. [Online]. Available: \url{https://data.europa.eu/eli/reg/2016/679/oj}
\BIBentrySTDinterwordspacing

\bibitem{Longo2024}
L.~Longo, M.~Brcic, F.~Cabitza, J.~Choi, R.~Confalonieri, J.~D. Ser, R.~Guidotti, Y.~Hayashi, F.~Herrera, A.~Holzinger, R.~Jiang, H.~Khosravi, F.~Lecue, G.~Malgieri, A.~Páez, W.~Samek, J.~Schneider, T.~Speith, and S.~Stumpf, ``Explainable artificial intelligence ({XAI}) 2.0: A manifesto of open challenges and interdisciplinary research directions,'' \emph{Information Fusion}, vol. 106, p. 102301, 2024.

\bibitem{Ali2023}
S.~Ali, T.~Abuhmed, S.~El-Sappagh, K.~Muhammad, J.~M. Alonso-Moral, R.~Confalonieri, R.~Guidotti, J.~{Del Ser}, N.~Díaz-Rodríguez, and F.~Herrera, ``Explainable artificial intelligence ({XAI}): What we know and what is left to attain trustworthy artificial intelligence,'' \emph{Information Fusion}, vol.~99, p. 101805, 2023.

\bibitem{ahmad2018interpretable}
M.~A. Ahmad, C.~Eckert, and A.~Teredesai, ``Interpretable machine learning in healthcare,'' in \emph{Proceedings of the 2018 ACM international conference on bioinformatics, computational biology, and health informatics}, 2018, pp. 559--560.

\bibitem{Maddigan2022}
P.~Maddigan and T.~Susnjak, ``Forecasting patient demand at urgent care clinics using machine learning,'' \emph{arXiv preprint arXiv:2205.13067}, 2022.

\bibitem{GPMaL}
A.~Lensen, B.~Xue, and M.~Zhang, ``Can genetic programming do manifold learning too?'' in \emph{Proceedings of the European Conference on Genetic Programming (EuroGP), Lecture Notes in Computer Science}.\hskip 1em plus 0.5em minus 0.4em\relax Springer International Publishing, 2019, vol. 11451, pp. 114--130.

\bibitem{GPMaLMO}
A.~Lensen, M.~Zhang, and B.~Xue, ``Multi-objective genetic programming for manifold learning: balancing quality and dimensionality,'' \emph{Genetic Programming and Evolvable Machines}, vol.~21, pp. 399--431, 2020.

\bibitem{review}
Y.~Mei, Q.~Chen, A.~Lensen, B.~Xue, and M.~Zhang, ``Explainable artificial intelligence by genetic programming: A survey,'' \emph{{IEEE} Transactions on Evolutionary Computation}, vol.~27, no.~3, pp. 621--641, 2023.

\bibitem{Maddigan2024}
P.~Maddigan, A.~Lensen, and B.~Xue, ``Explaining genetic programming trees using large language models,'' \emph{arXiv preprint arXiv:2403.03397}, 2024.

\bibitem{Carvalho2019}
D.~V. Carvalho, E.~M. Pereira, and J.~S. Cardoso, ``Machine learning interpretability: A survey on methods and metrics,'' \emph{Electronics}, vol.~8, no.~8, 2019.

\bibitem{Singh2024}
C.~Singh, J.~P. Inala, M.~Galley, R.~Caruana, and J.~Gao, ``Rethinking interpretability in the era of large language models,'' \emph{arXiv preprint arXiv:2402.01761}, 2024.

\bibitem{Molnar2020}
C.~Molnar, G.~Casalicchio, and B.~Bischl, ``Interpretable machine learning -- a brief history, state-of-the-art and challenges,'' in \emph{ECML PKDD 2020 Workshops}.\hskip 1em plus 0.5em minus 0.4em\relax Cham: Springer International Publishing, 2020, pp. 417--431.

\bibitem{Domingos2012}
P.~Domingos, ``A few useful things to know about machine learning,'' \emph{Communications of the ACM}, vol.~55, no.~10, pp. 78--87, 2012.

\bibitem{10007067}
J.~Huang, W.~Qian, C.-M. Vong, W.~Ding, W.~Shu, and Q.~Huang, ``Multi-label feature selection via label enhancement and analytic hierarchy process,'' \emph{IEEE Transactions on Emerging Topics in Computational Intelligence}, vol.~7, no.~5, pp. 1377--1393, 2023.

\bibitem{Zebari2020}
R.~Zebari, A.~Abdulazeez, D.~Zeebaree, D.~Zebari, and J.~Saeed, ``A comprehensive review of dimensionality reduction techniques for feature selection and feature extraction,'' \emph{Journal of Applied Science and Technology Trends}, vol.~1, no.~2, pp. 56 -- 70, May 2020.

\bibitem{van2009dimensionality}
L.~Van Der~Maaten, E.~O. Postma, H.~J. van~den Herik \emph{et~al.}, ``Dimensionality reduction: A comparative review,'' \emph{Journal of Machine Learning Research}, vol.~10, no. 66-71, p.~13, 2009.

\bibitem{mcinnes2020umap}
L.~McInnes, J.~Healy, and J.~Melville, ``Umap: Uniform manifold approximation and projection for dimension reduction,'' 2020.

\bibitem{auto}
G.~E. Hinton and R.~Salakhutdinov, ``Reducing the dimensionality of data with neural networks,'' \emph{Science}, vol. 313, pp. 504 -- 507, 2006.

\bibitem{t-SNE}
L.~van~der Maaten and G.~E. Hinton, ``Visualizing data using t-sne,'' \emph{Journal of Machine Learning Research}, vol.~9, pp. 2579--2605, 2008.

\bibitem{distill}
M.~Wattenberg, F.~Viégas, and I.~Johnson, ``How to use t-sne effectively,'' \emph{Distill}, 2016.

\bibitem{9881543}
J.~Dong, J.~Zhong, W.-N. Chen, and J.~Zhang, ``An efficient federated genetic programming framework for symbolic regression,'' \emph{IEEE Transactions on Emerging Topics in Computational Intelligence}, vol.~7, no.~3, pp. 858--871, 2023.

\bibitem{poli2008field}
R.~Poli, W.~B. Langdon, and N.~F. McPhee, \emph{A Field Guide to Genetic Programming}.\hskip 1em plus 0.5em minus 0.4em\relax lulu.com, 2008.

\bibitem{10466603}
B.~Al-Helali, Q.~Chen, B.~Xue, and M.~Zhang, ``Genetic programming for feature selection based on feature removal impact in high-dimensional symbolic regression,'' \emph{IEEE Transactions on Emerging Topics in Computational Intelligence}, pp. 1--14, 2024, early Access.

\bibitem{coello2006evolutionary}
C.~A.~C. Coello, ``Evolutionary multi-objective optimization: a historical view of the field,'' \emph{{IEEE} Comput. Intell. Mag.}, vol.~1, no.~1, pp. 28--36, 2006.

\bibitem{zhang2007moead}
Q.~Zhang and H.~Li, ``{MOEA/D:} {A} multiobjective evolutionary algorithm based on decomposition,'' \emph{{IEEE} Trans. Evol. Comput.}, vol.~11, no.~6, pp. 712--731, 2007.

\bibitem{expression}
M.~Kommenda, G.~Kronberger, M.~Affenzeller, S.~M. Winkler, and B.~Burlacu, \emph{Evolving Simple Symbolic Regression Models by Multi-Objective Genetic Programming}.\hskip 1em plus 0.5em minus 0.4em\relax Springer International Publishing, 2016, ch. Genetic Programming Theory and Practice XIII, pp. 1--19.

\bibitem{time}
A.~S. Sambo, R.~M.~A. Azad, Y.~Kovalchuk, V.~P. Indramohan, and H.~Shah, ``Time control or size control? reducing complexity and improving the accuracy of genetic programming models,'' 2020.

\bibitem{Le2016}
N.~Le, H.~N. Xuan, A.~Brabazon, and T.~P. Thi, ``Complexity measures in genetic programming learning: A brief review,'' in \emph{2016 IEEE Congress on Evolutionary Computation (CEC)}, 2016, pp. 2409--2416.

\bibitem{chadalawada}
J.~Chadalawada, V.~Havlicek, and V.~Babovic, ``A genetic programming approach to system identification of rainfall-runoff models,'' \emph{Water Resour Manage}, vol.~31, p. 3975–3992, 2017.

\bibitem{campobello}
G.~Campobello, D.~Dell’Aquila, M.~Russo, and A.~Segreto, ``Neuro-genetic programming for multigenre classification of music content,'' \emph{Applied Soft Computing}, vol.~94, p. 106488, 2020.

\bibitem{Tang}
Y.~Tang, H.~Jia, and N.~Verma, ``Reducing energy of approximate feature extraction in heterogeneous architectures for sensor inference via energy-aware genetic programming,'' \emph{IEEE Transactions on Circuits and Systems I: Regular Papers}, vol.~67, no.~5, pp. 1576--1587, 2020.

\bibitem{Soule1998EffectsOC}
T.~Soule and J.~A. Foster, ``Effects of code growth and parsimony pressure on populations in genetic programming,'' \emph{Evolutionary Computation}, vol.~6, pp. 293--309, 1998.

\bibitem{Smits2005ParetoFrontEI}
G.~F. Smits and M.~Kotanchek, \emph{Pareto-Front Exploitation in Symbolic Regression}.\hskip 1em plus 0.5em minus 0.4em\relax Springer US, 2005, pp. 283--299.

\bibitem{tik}
J.Ni and P.Rockett, ``Tikhonov regularization as a complexity measure in multiobjective genetic programming,'' \emph{IEEE Transactions on Evolutionary Computation}, vol.~19, no.~2, pp. 157--166, 2015.

\bibitem{UCI}
\BIBentryALTinterwordspacing
D.~Dheeru and E.~Karra~Taniskidou, ``{UCI} machine learning repository,'' 2017. [Online]. Available: \url{http://archive.ics.uci.edu/ml}
\BIBentrySTDinterwordspacing

\bibitem{Nene1996}
S.~A. Nene, S.~K. Nayar, and H.~Murase, ``Columbia object image library (coil-20),'' Columbia University, Tech. Rep., 1996.

\end{thebibliography}

\end{document}